\documentclass[letterpaper]{article} 
\usepackage{aaai23}  
\usepackage{times}  
\usepackage{helvet}  
\usepackage{courier}  
\usepackage[hyphens]{url}  
\usepackage{graphicx} 
\usepackage{natbib}  
\usepackage{caption} 
\usepackage{algorithm}
\usepackage{algorithmic}

\usepackage{graphicx}

\usepackage{tikz}
\usepackage{comment}
\usepackage{amsmath,amssymb} 
\usepackage{color}

\usepackage[accsupp]{axessibility}  

\usepackage{multirow}
\usepackage{adjustbox}
\usepackage{subcaption}
\usepackage[labelfont=bf]{caption}
\usepackage{setspace}
\usepackage{overpic}
\usepackage{bm}
\usepackage{colortbl}
\usepackage{diagbox}
\usepackage{mathtools}
\definecolor{maroon}{cmyk}{0,0.87,0.68,0.32}

\usepackage{multirow}
\usepackage{booktabs}
\usepackage{array}
\newcommand{\PreserveBackslash}[1]{\let\temp=\\#1\let\\=\temp}
\newcolumntype{C}[1]{>{\PreserveBackslash\centering}p{#1}}
\usepackage{tikz}

\usepackage{bbding}
\usepackage{overpic}
\usepackage{makecell}
\usepackage[colorlinks,linkcolor=blue]{hyperref}
\def\wrt{\textit{w.r.t}}
\def\eg{\textit{e.g.}}
\def\ie{\textit{i.e.}}

\usepackage{newfloat}
\usepackage{listings}
\DeclareCaptionStyle{ruled}{labelfont=normalfont,labelsep=colon,strut=off} 
\floatstyle{ruled}
\newfloat{listing}{tb}{lst}{}
\floatname{listing}{Listing}
\title{PolarFormer: Multi-camera 3D Object Detection with
Polar Transformer}
\author{
    Yanqin Jiang\textsuperscript{\rm 1,4}\thanks {Work done while at Fudan University.},
    Li Zhang\textsuperscript{\rm 2}\thanks{Li Zhang (lizhangfd@fudan.edu.cn) is the corresponding author.
    },
    Zhenwei Miao\textsuperscript{\rm 5},
    Xiatian Zhu\textsuperscript{\rm 6},
    Jin Gao\textsuperscript{\rm 1,4},\\
    Weimin Hu\textsuperscript{\rm 1,4,7},
    Yu-Gang Jiang\textsuperscript{\rm 3}
}

\affiliations{
    \textsuperscript{\rm 1}NLPR, Institute of Automation, Chinese Academy of Sciences,\\
    \textsuperscript{\rm 2}School of Data Science, Fudan University,
    \textsuperscript{\rm 3}School of Computer Science, Fudan University, \\
    \textsuperscript{\rm 4}School of Artificial Intelligence, University of Chinese Academy of Sciences,\\
    \textsuperscript{\rm 5}Alibaba DAMO Academy,
    \textsuperscript{\rm 6}Surrey Institute for People-Centred Artificial Intelligence, CVSSP, University of Surrey,\\
    \textsuperscript{\rm 7}School of Information Science and Technology, ShanghaiTech University \\
    \vspace{.5em} 
    \url{https://github.com/fudan-zvg/PolarFormer}
}

\usepackage{bibentry}

\begin{document}

\maketitle

\begin{abstract}
3D object detection in autonomous driving aims to reason ``what'' and ``where'' the objects of interest present
in a 3D world.
Following the conventional wisdom of previous 2D object detection,
existing methods often adopt the canonical Cartesian coordinate system with perpendicular axis.
However, we conjugate that this does not fit the nature of the ego car's perspective, as each onboard camera perceives
the world in shape of wedge 
intrinsic to the imaging geometry with radical (non-perpendicular) axis.
Hence, in this paper we advocate the exploitation of the Polar coordinate system and propose a new Polar Transformer ({\bf\em PolarFormer})
for more accurate 3D object detection in the bird's-eye-view (BEV)
taking as input only multi-camera 2D images.
Specifically, we design a cross-attention based Polar detection head without
restriction to the shape of input structure to deal with irregular Polar grids.
For tackling the unconstrained object scale variations along Polar's distance dimension, we further introduce a multi-scale Polar representation learning strategy.
As a result, our model can make best use of the Polar representation rasterized via attending to the corresponding image observation
in a sequence-to-sequence fashion subject to the geometric constraints.
Thorough experiments on the nuScenes dataset  demonstrate that our PolarFormer outperforms significantly state-of-the-art 3D object detection alternatives.
\end{abstract}

\newcommand\blfootnote[1]{%
\begingroup 
\renewcommand\thefootnote{}\footnote{#1}%
\addtocounter{footnote}{-1}%
\endgroup 
}
\section{Introduction}

\begin{figure*}[h]
    \centering
    \includegraphics[width=1\textwidth]{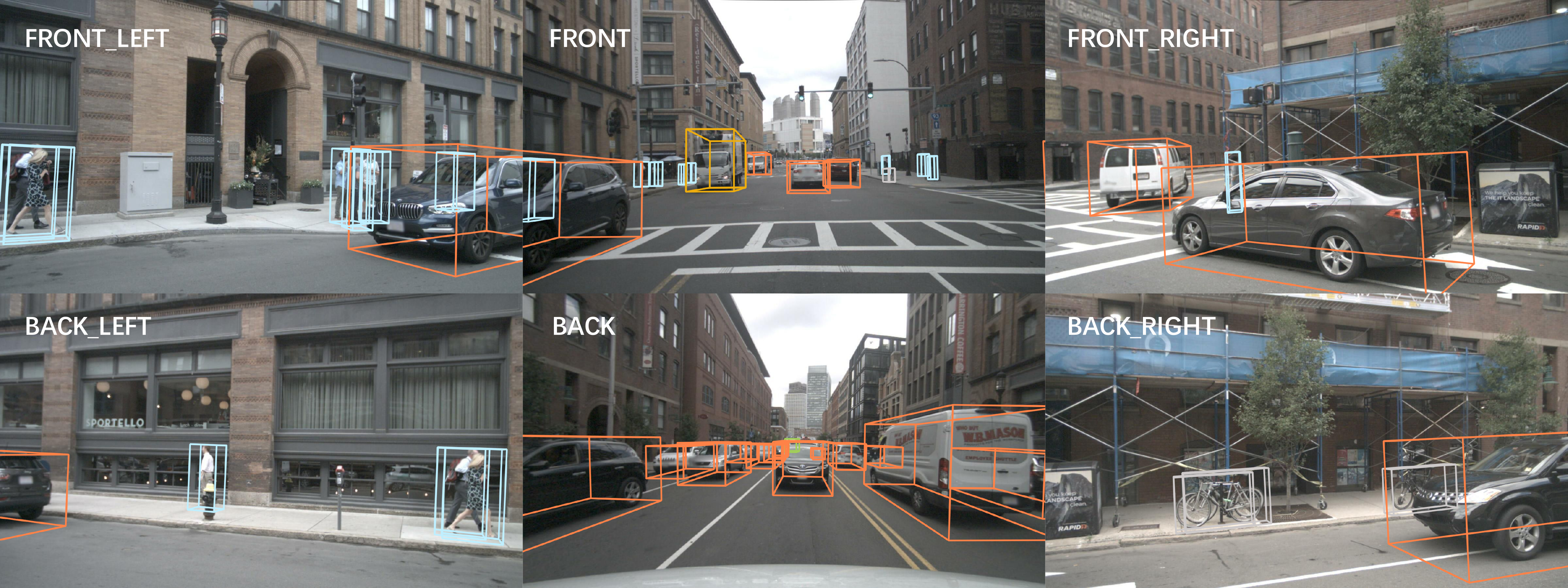}
    \caption{
    Taking multi-camera images as input, the proposed \texttt{PolarFormer} model is designed particularly 
    for accurate 3D object detection
    in the Polar coordinate system.
    }
    \label{fig:teaser}
\end{figure*}

3D object detection is an enabling capability of autonomous driving in unconstrained real-world scenes \cite{wang2022detr3d,wang2021fcos3d}.
It aims to predict the location, dimension and orientation of individual objects of interest in a 3D world.
Despite favourable cost advantages, multi-camera based 3D object detection~\cite{wang2021fcos3d,wang2022probabilistic,wang2022detr3d,zhou2019objects} remains particularly challenging.
To obtain 3D representation, 
dense depth estimation is often leveraged \cite{philion2020lift},
which however is not only expensive in computation
but error prone.
To bypass depth inference, more recent methods \cite{wang2022detr3d,li2022bevformer} exploit 
query-based 2D detection~\cite{carion2020end}
to learn a set of sparse and virtual embedding for multi-camera 3D object detection, yet incapable of effectively modeling the geometry structure among objects.
Typically, the canonical Cartesian coordinate system with {\em perpendicular} axis is adopted in
either 2D \cite{zhou2019objects,wang2021fcos3d} or 3D \cite{wang2022detr3d,li2022bevformer} space.
This is largely restricted by convolution based models used.
In contrast, the physical world perceived under each camera {\em in the ego car's perspective} is in shape of wedge intrinsic to the camera imaging geometry with radical 
{\em non-perpendicular} axis (Figure \ref{fig:polarcardi}).
Bearing this imaging property in mind, we conjugate that the Polar coordinate system should be more appropriate and natural than the often adopted Cartesian counterpart
for 3D object detection.
Indeed, the Polar coordinate has been exploited in a few LiDAR-based 3D perception methods \cite{zhang2020polarnet,bewley2020range,rapoport2021s,zhu2021cylindrical}.
However, they are limited algorithmically due to the adoption of convolutional networks restricted to rectangular grid structure and local receptive fields.

Motivated by the aforementioned insights, in this work a novel {\bf\em Polar Transformer} (PolarFormer) model for multi-camera 3D object detection in a Polar coordinate system is introduced (Figure~\ref{fig:learning_targets}).
Specifically, we first learn the representation of polar rays corresponding to image regions in a sequence-to-sequence cross-attention formulation.
Then we rasterize a BEV Polar representation consisting of a set of Polar rays evenly distributed around 360 degrees.
To deal with irregular Polar grids as suffered by conventional LiDAR based solutions \cite{zhang2020polarnet,bewley2020range,rapoport2021s,zhu2021cylindrical}, 
we propose a cross-attention based decoder head design without
restriction to the shape of input structure.
For tackling the challenge of unconstrained object scale variation along Polar's distance dimension, we resort to a multi-scale Polar BEV representation learning strategy.

The {\bf contributions} of this work are summarized as follows: 
\textbf{(I)}
We propose a new {\em Polar Transformer} (PolarFormer) model for multi-camera 3D object detection in the Polar coordinate system.
\textbf{(II)}
This is achieved based on two Polar-tailored designs:
A cross-attention based decoder design
for dealing with the irregular Polar girds,
and a multi-scale Polar representation learning strategy
for handling the unconstrained object scale variations over Polar's distance dimension.
\textbf{(III)}
Extensive experiments on the nuScenes dataset show that our PolarFormer achieves leading performance for camera-based 3D object detection (Figure \ref{fig:teaser}).
\section{Related work}

\paragraph{Monocular/multi-camera 3D object detection}
Image-based 3D object detection aims to estimate the object location, dimension and orientation in the 3D space alongside its category given only image input.
To solve this ill-posed problem, 
\cite{zhou2019objects,wang2021fcos3d} naively build their detection pipelines by augmenting 2D detectors~\cite{zhou2019objects,tian2019fcos} in a data driven fashion.
PGD~\cite{wang2022probabilistic} further captures the uncertainty and models the relationship of different objects by utilizing geometric prior. 
Contrast to the above image-only methods, depth-based methods~\cite{xu2018multi,ding2020learning,wang2019pseudo,you2019pseudo,ma2019accurate,reading2021categorical} use depth cues as 3D information to mitigate the naturally ill-posed problem.
Recently, multi-camera-based 3D object detection emerges. DETR3D~\cite{wang2022detr3d} considers detecting objects across all cameras collectively. 
It learns a set of sparse and virtual query embedding, without explicitly building the geometry structure among objects/queries. 
~\cite{li2022bevformer} considers detecting objects in BEV,
performing end-to-end object detection via object queries.
Note that multi-camera setting uses the same amount of training data as the monocular pipelines.
Both multi-camera and monocular paradigms share the same evaluation metrics.
\paragraph{Bird's-eye-view (BEV) representation}
Recently there is a surge of interest in transforming the monocular or multi-view images from ego car cameras into the bird's-eye-view coordinate~\cite{roddick2018orthographic,philion2020lift,li2021hdmapnet,roddick2020predicting,reading2021categorical,saha2021translating} followed by specific optimization tasks (\eg,~3D object detection, semantic segmentation).
A natural solution~\cite{philion2020lift,hu2021fiery} is to 
learn the BEV representation by leveraging the pixel-level dense depth estimation. This however is error-prone due to lacking ground-truth supervision.
Another line of research aims to bypass the depth prediction and directly leverage a Transformer~\cite{chitta2021neat,can2021structured,saha2021translating} or a FC layer~\cite{li2021hdmapnet,roddick2020predicting,yang2021projecting} to learn the transformation from camera inputs to the BEV coordinate.
A similar attempt as ours is conducted in \cite{saha2021translating} but limited in a couple of aspects:
{\bf (i)}
It is restricted to monocular input for a straightforward 2D segmentation task while we consider multiple cameras collectively for more challenging 3D object detection;
{\bf (ii)}
We uniquely provide a solid multi-scale Polar BEV transformation to tackle the unconstrained object scale variations and followed by a jointly optimized cross-attention based Polar head.
\paragraph{3D object detection in Polar coordinate}
3D object detection in the Polar or Polar-like coordinate system
has been attempted in LiDAR-based perception methods.
For example, CyliNet~\cite{zhu2021cylindrical} introduces range-based guidance for extracting Polar-consistent features.
In particular, it adapts a Cartesian heatmap to a Polar version for object classification, whilst learning relative heading angles and velocities. However, CyliNet still lags clearly behind the Cartesian counterpart. 
Recently, PolarSteam~\cite{chen2021polarstream} designs a learnable sampling module for relieving object distortion in Polar coordinate and uses range-stratified convolution and normalization for flexibly extracting the features over different ranges. 
Limited by the convolution based network, it remains inferior 
despite of these special designs.
In contrast to all these works, we resort to the cross-attention mechanism, tackling the challenges of object scale variance and appearance distortion in the Polar coordinate principally.
\section{Method}
\begin{figure*}[ht]
    \centering
    \begin{overpic}[width=1.02\textwidth]
    {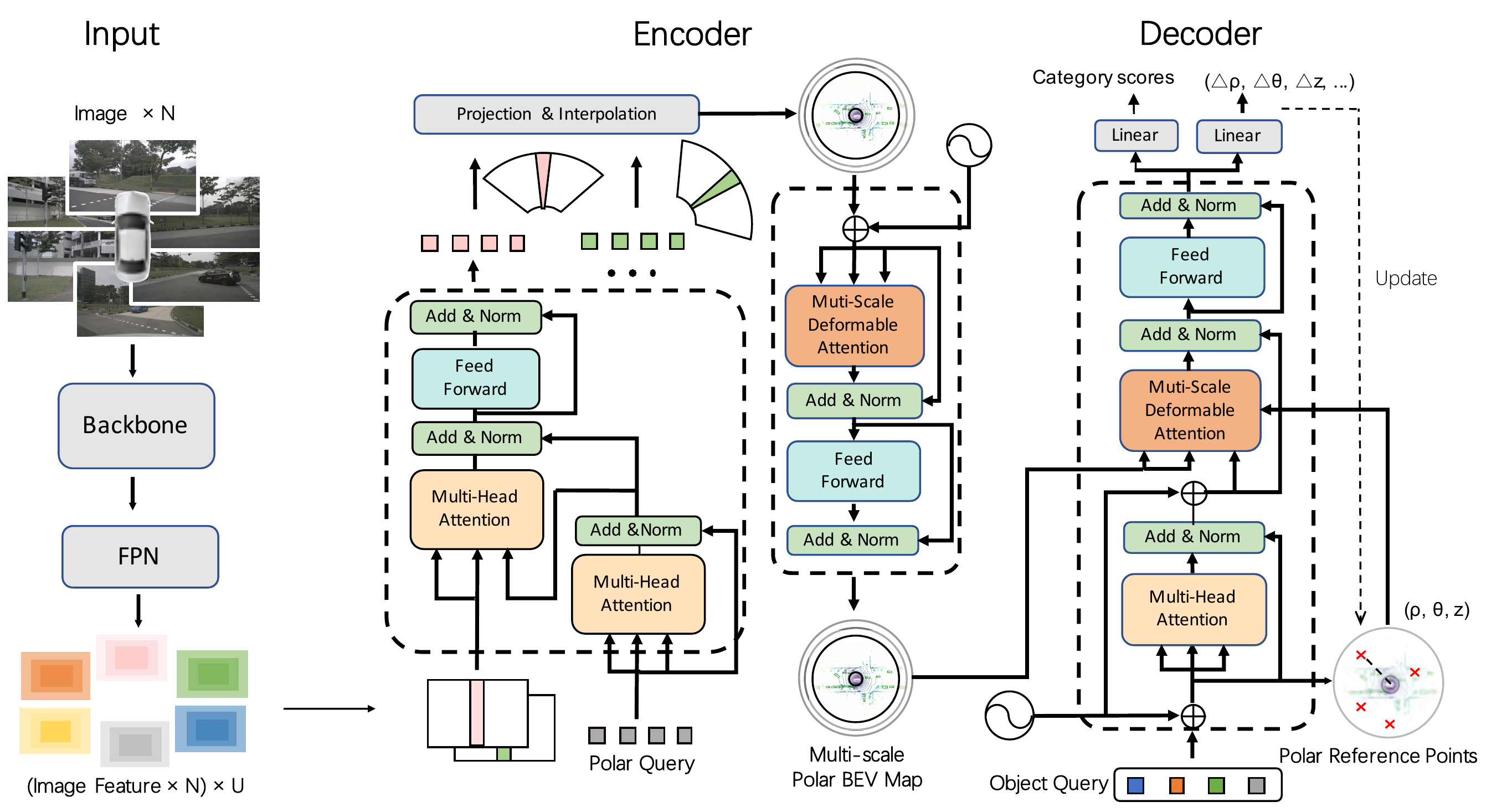}
    \put(43.0,31.0){\scriptsize \textbf{(a)}}
    \put(40.0,29.5){\scriptsize Cross-plane}
    \put(41.0,28.0){\scriptsize encoder}
    \put(27,48.0){\scriptsize \textbf{(b)} Polar alignment module}
    \put(62.0, 13.5){\scriptsize \textbf{(c)}}
    \put(59.5, 12.0){\scriptsize Polar BEV}
    \put(61.0, 10.5){\scriptsize encoder}
    \end{overpic}
    \caption{Schematic illustration of our proposed {\em PolarFormer} for multi-camera 3D object detection. 
    For each image captured by any camera view, our model first extracts the feature maps at multiple spatial scales. 
    Given such a feature map, the cross-plane encoder \textbf{(a)} then transforms all the feature columns to a set of Polar rays in a sequence-to-sequence manner via polar queries based cross-attention. The polar rays from all the cameras are subsequently processed by a Polar alignment module \textbf{(b)} to generate a structured multi-scale Polar BEV map, followed by further enhancement via interactions among different scales using a Polar BEV encoder \textbf{(c)}.
    At last, a specially designed Polar Head decodes multi-scale Polar BEV features for making final predictions in the Polar coordinate.}
    \label{fig:pipeline}
\end{figure*}
\begin{figure*}[t]
\centering
\begin{minipage}[t]{0.445\textwidth}
\centering
\includegraphics[width=1\textwidth, trim={0 0.285cm 0 0}]{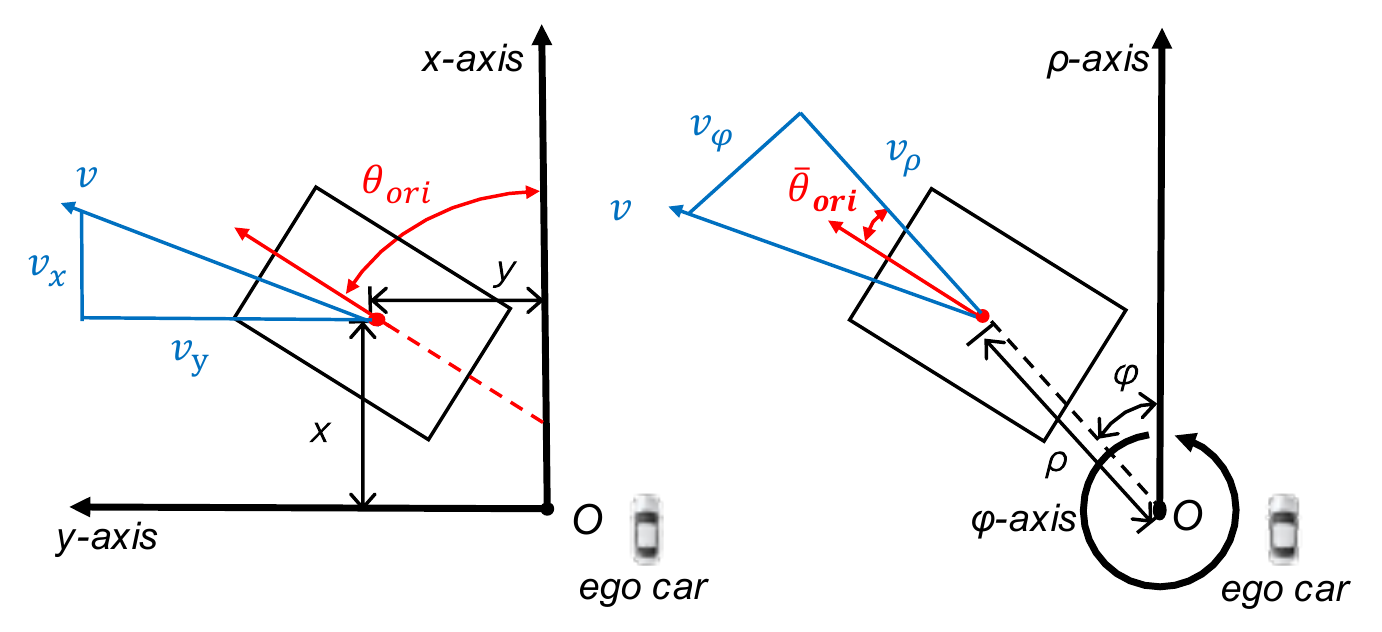}
\caption{Cartesian and Polar coordinates.}
\label{fig:learning_targets}
\end{minipage}
\begin{minipage}[t]{0.545\textwidth}
\centering
\includegraphics[width=0.5\textwidth,trim={0cm 0cm 0 0cm}]{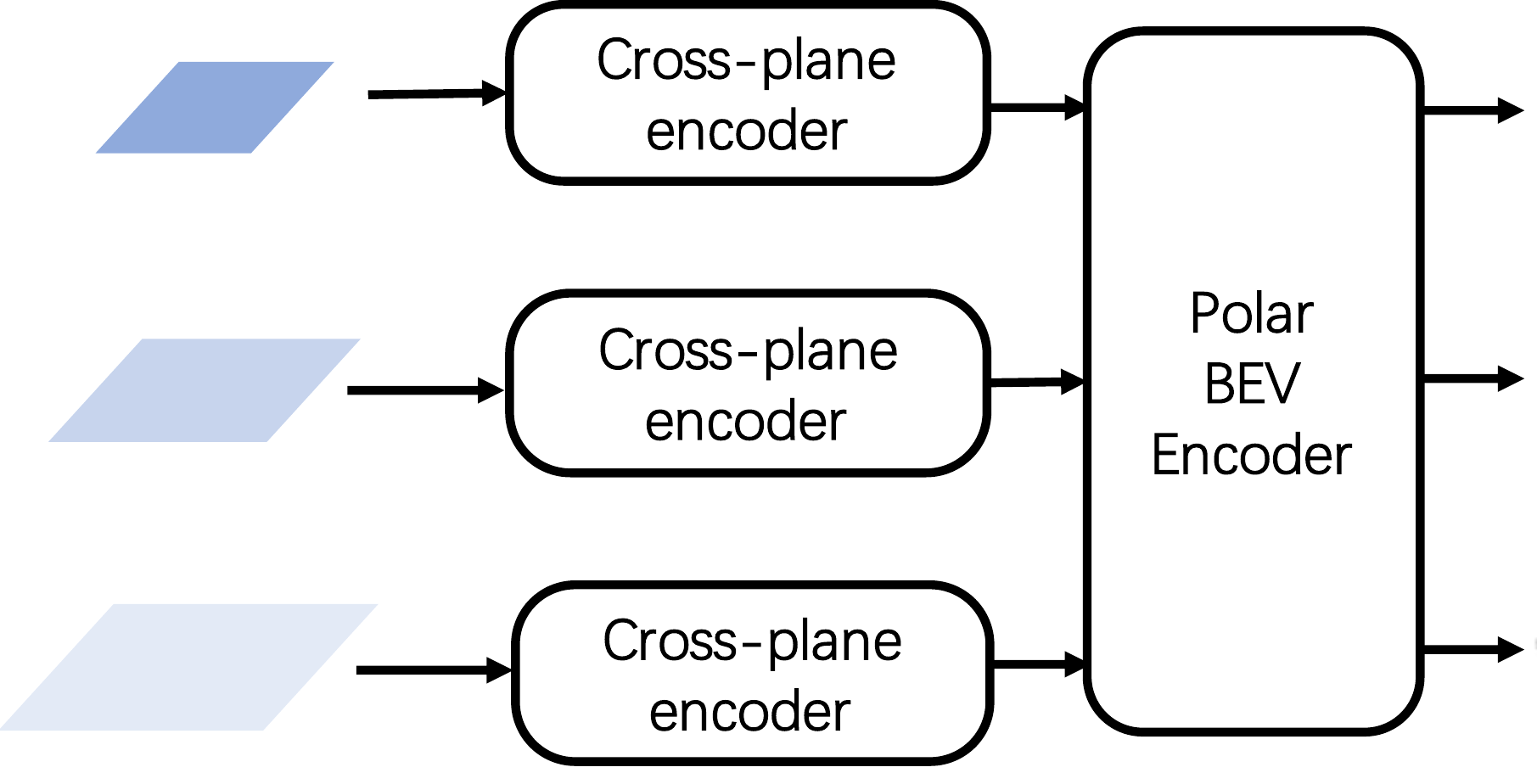}
\hspace{-0.75cm}
\includegraphics[height=2.225cm, width=0.3\textwidth,trim={-4cm -2cm 0 0cm}]{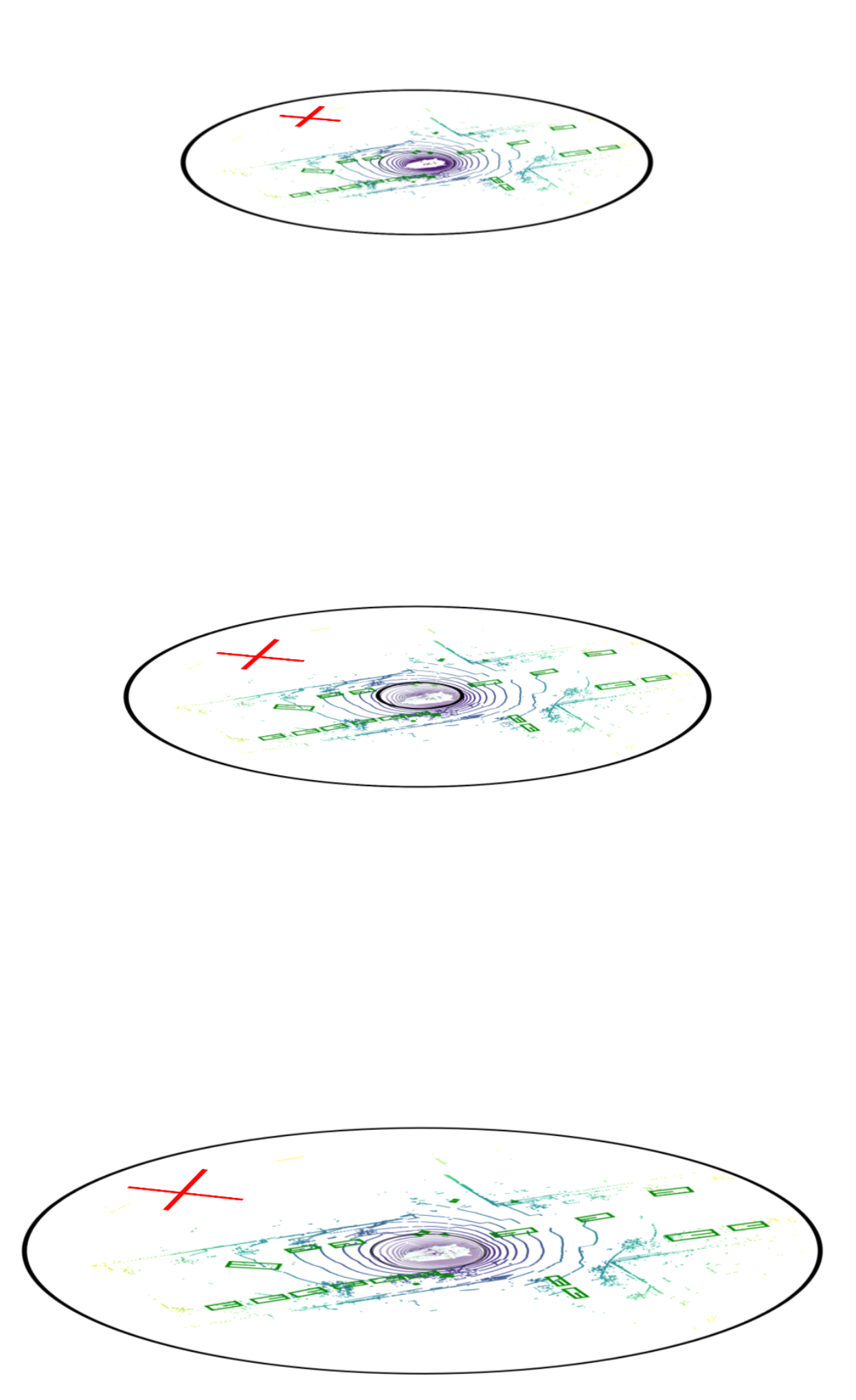} 
\hspace{-0.15cm}
\includegraphics[width=0.225\textwidth,trim={0cm 0cm 0 0cm}]{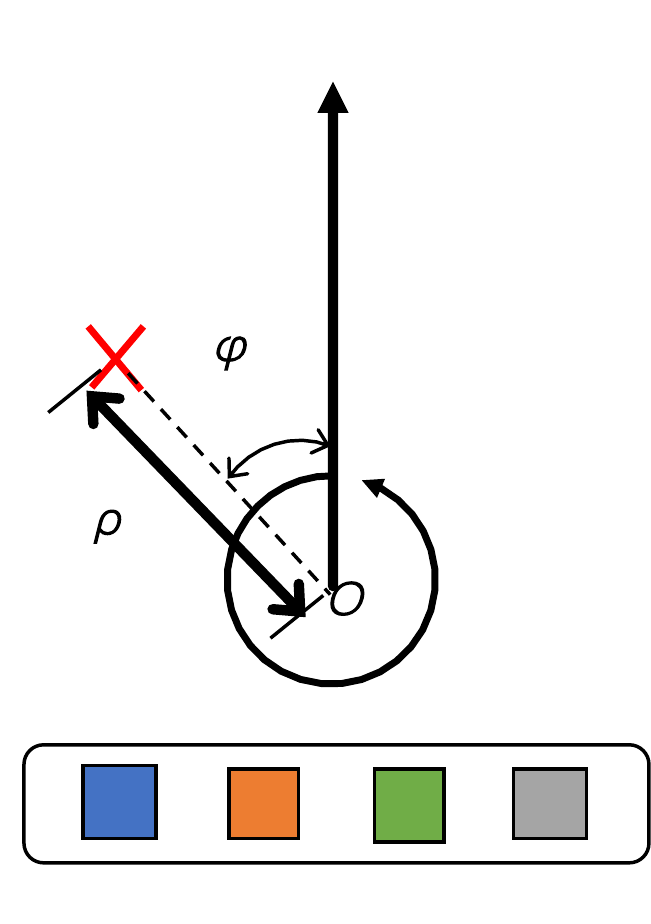}
\caption{Multi-scale Polar BEV maps.}
\label{fig:multi-scale}
\end{minipage}
\end{figure*}

In 3D object detection task, we are given a set of $N$ monocular views $\{\mathbf{I_n}, \mathbf{\Pi_n}, \mathbf{E_n}\}_{n=1}^N$ consisting of input images $\mathbf{I_n} \in \mathbb{R}^{H\times W\times 3}$, camera intrinsics $\mathbf{\Pi_n} \in \mathbb{R}^{3\times3}$ and camera extrinsics $\mathbf{E_n} \in \mathbb{R}^{4\times4}$. 
The objective of our {\em Polar Transformer} (PolarFormer) is to learn an effective 
BEV Polar representation 
from multiple camera views for facilitating the prediction of object locations, dimensions, orientations and velocities in the Polar coordinate system.
PolarFormer consists of the following components.
A {\em cross-plane encoder} first
produces a multi-scale feature representation of
each input image, characterized by a cross-plane attention mechanism in which Polar queries attend to input images to generate 3D features in BEV. 
A {\em Polar alignment module} then aggregates Polar rays from multiple camera views to generate a structured Polar map. Further, a {\em BEV Polar encoder} enhances the Polar features with multi-scale feature interaction. Finally, a {\em Polar detection head} decodes the Polar map and predicts the objects in the Polar coordinate system. 
For tackling the unconstrained object scale variation with multi-granularity of details,
we consider a multi-scale BEV Polar representation structure. 
As shown in Figure \ref{fig:multi-scale}, image features with different scales have unique cross-plane encoders
and interact with each other in a shared Polar BEV encoder. 
Multi-scale Polar BEV maps are then queried by Polar decoder head.
An overall architecture of PolarFormer is depicted in Figure \ref{fig:pipeline}.

\subsection{Cross-plane encoder}

The goal of cross-plane encoder is to associate an image with BEV Polar rays. According to the geometric model of camera, for any camera coordinate $ (x^{(C)}, y^{(C)}, z^{(C)})\in \mathbb{R}^3$, the transformation to image coordinate $ (x^{(I)}, y^{(I)})\in \mathbb{R}^2$could be described as:
\begin{equation}\label{eq:cam2img}
    s\begin{bmatrix}
    x^{(I)}\\
    y^{(I)}\\
    1
    \end{bmatrix} =
    \begin{bmatrix}
    f_x & 0 & u_0 \\
    0 & f_y & v_0 \\
    0 & 0 & 1
    \end{bmatrix}
    \begin{bmatrix}
    x^{(C)} \\
    y^{(C)} \\
    z^{(C)} 
    \end{bmatrix},
\end{equation}
where $f_x$, $f_y$, $u_0$ and $v_0$ are camera intrinsic parameters in $\mathbf{\Pi}$, $x^{(C)}$, $y^{(C)}$ and $z^{(C)}$ are the horizontal, vertical, depth coordinate respectively. $s$ is the scale factor. For any BEV Polar coordinate $(\rho^{(P)}, \phi^{(P)})$, we have:
\begin{equation}\label{eq:amizuth}
    \phi^{(P)} = \arctan \frac{x^{(C)}}{z^{(C)}} = \arctan \frac{x^{(I)}-u_0}{f_x},
\end{equation}
\begin{equation}\label{eq:radius}
\begin{split}
    \rho^{(P)} = \sqrt{(x^{(C)})^2 + (z^{(C)})^2} 
         =z^{(C)}\sqrt{(\frac{x^{(I)}-u_0}{f_x})^2+1}.
\end{split}
\end{equation}
Eq. \eqref{eq:amizuth} suggests that the azimuth $\phi^{(P)}$ is irrelevant to the vertical value of image coordinate. It is hence natural to build a one-to-one relationship between Polar rays and image columns \cite{saha2021translating}. However, we need object depth $z^{(C)}$ to compute radius $\rho^{(P)}$, the estimation of which is ill-posed. Instead of explicit depth estimation, we leverage 
the attention mechanism~\cite{vaswani2017attention} to model the relationship between pixels along the image column and positions along the Polar ray.  

Let $\mathbf{f}_{n,u,w} \in \mathbb{R}^{H_u\times C}$ represent the image column from $n$th camera, $u$th scale and $w$th column, and
$\mathbf{\dot{p}}_{n,u,w} \in \mathbb{R}^{R_u \times C}$ denote the corresponding {\em Polar ray} query we introduce, where $H$ and $R$ are the image feature map's height and Polar map's range. We formulate cross-plane attention as:
\begin{equation}
  \begin{split}
  \mathbf{p}_{n,u,w} &= \mathrm{MultiHead}(\mathbf{\dot{p}}_{n,u,w},\mathbf{f}_{n,u,w},\mathbf{f}_{n,u,w}) \\
  &= \mathrm{Concat}(\mathrm{head_1}, \dots, \mathrm{head_h})\mathbf{W}^O_u,
  \end{split}
\end{equation}
where
\begin{equation}
  \begin{split}
  \mathrm{head_i} \!=\! \mathrm{Attention}(\mathbf{\dot{p}}_{n,u,w}\mathbf{W}_{i,u}^Q, \mathbf{f}_{n,u,w}\mathbf{W}_{i,u}^K, \mathbf{f}_{n,u,w}\mathbf{W}_{i,u}^V),
  \end{split}
\end{equation}
where $\mathbf{W}_{i,u}^Q \in \mathbb{R}^{d_{model} \times d_q}$, $\mathbf{W}_{i,u}^K \in \mathbb{R}^{d_{model} \times d_k}$, $\mathbf{W}_{i,u}^V \in \mathbb{R}^{d_{model} \times d_v}$, $\mathbf{W}^O_u \in \mathbb{R}^{hd_{model} \times d_k}$
are the projection parameters, $d_q=d_k=d_v=d_{model}/h$, $d$ is the feature dimension and $h$ is the number of heads. 

Stacking the Polar ray features $\mathbf{p}_{n,u,w} \in \mathbb{R}^{R_u \times C}$ along azimuth axis, we obtain the Polar feature map (\ie, {\em BEV Polar representation}) $\mathbf{P}_{n,u}$ for $n$th camera and $u$th scale as:
\begin{equation}
    \mathbf{P}_{n,u}\!=\!\mathrm{Stack}([\mathbf{p}_{n,u,1}, \dots,\mathbf{p}_{n,u,W_u}], \mathrm{dim}\!=\!1) \!\in\! \mathbb{R}^{R_u \!\times\! W_u \!\times\! C },
\end{equation}
where $W_u$ denotes the azimuth dimension.
This sequence-to-sequence cross-attention-based encoder can encode geometric imaging prior and implicitly learn a proxy for depth efficiently. Next, we show how to integrate independent Polar rays from multiple cameras into a coherent and structured Polar BEV map. 

\subsection{Polar alignment across multiple cameras}
Our Polar alignment module transforms Polar rays from different camera coordinates to a shared world coordinate. Taking multi-view Polar feature maps $\{\mathbf{P}_{n,u}\}_{n=1}^N$ and camera matrix $\{\mathbf{\Pi}_n,\mathbf{E}_n\}_{n=1}^N$as inputs, it produces a coherent BEV Polar map $\mathbf{G}_u \in \mathbb{R}^{\mathcal{R}_u\times\mathcal{N}_u\times \mathcal{C}}$, covering all camera views, where $\mathcal{R}_u$, $\mathcal{N}_u$ and $\mathcal{C}$ are the dimensions of radius, azimuth and feature. 
Concretely, it first generates a set of 3D points in the cylindrical coordinate uniformly, denoted by $\mathcal{G}^{(P)}=\{(\rho^{(P)}_{i},\phi^{(P)}_{j},z^{(P)}_{k})|i=1,\dots,\mathcal{R}_u;j=1,\dots,\mathcal{N}_u; k=1,\dots,\mathcal{Z}_u\}$, where $\mathcal{Z}_u$ is the number of points along $z$ axis. Since cylindrical coordinate and Polar coordinate share radius and azimuth axis, their superscripts are both denoted with $P$. The points are then projected to the image plane of $n$th camera to retrieve the index of Polar ray, estimated by:
\begin{equation}\label{eq:alignment_proj}
    \begin{bmatrix}
    sx^{(I)}_{i,j,k,n} \\[5pt]
    sy^{(I)}_{i,j,k,n}\\[5pt]
    s \\[5pt]
    1
    \end{bmatrix} =
    \begin{bmatrix}
    \mathbf{\Pi}_n & 0 \\
    0 & 1
    \end{bmatrix}
    \mathbf{E}_n
    \begin{bmatrix}
    \rho^{(P)}_i\sin{\phi^{(P)}_j} \\[5pt]
    \rho^{(P)}_i\cos{\phi^{(P)}_j} \\[5pt]
    z^{(P)}_k \\[5pt]
    1
    \end{bmatrix},
\end{equation}
where $s$ is the scale factor. Coherent BEV Polar map of $u$th scale can be then generated by:
\begin{equation}
\label{eq:corherent_sampling}
\begin{split}
    &\mathbf{G}_u(\rho^{(P)}_i,\phi^{(P)}_j)
    =\frac{1}{\sum_{n=1}^N\sum_{k=1}^\mathcal{Z} \lambda_n(\rho^{(P)}_i,\phi^{(P)}_j,z^{(P)}_k)} \\ 
    & \cdot \sum_{n=1}^N\sum_{k=1}^\mathcal{Z}\lambda_n(\rho^{(P)}_i,\phi^{(P)}_j,z^{(P)}_k){\mathcal{B}(\mathbf{P}_{n,u},(\overline{x}^{(I)}_{i,j,k,n}, \overline{r}_{i,j,n}))},
\end{split}
\end{equation}
where $\lambda_n(\rho^{(P)}_i,\phi^{(P)}_j,z^{(P)}_k)$ is binary weighted factor indicating visibility in $n$th camera, $\mathcal{B}(\mathbf{P}, (x,y))$ denotes the bilinear sampling $\mathbf{P}$ at location $(x,y)$, $\overline{x}^{(I)}_{i,j,k,n}$ and  $\overline{r}_{i,j,n}$ denote the normalized Polar ray index and radius index. Note the radius $r$ is the distance between the point and the camera origin in BEV. Our Polar alignment module incorporates the features at different heights by generating the points along $z$ axis. 
As validated in Table \ref{table:coordinate}, learning Polar representation is superior over Cartesian coordinate due to  
minimal information loss and higher consistency with raw visual data.

\begin{table*}[!t]
\footnotesize
  \centering
\caption{
State-of-the-art comparison on nuScenes \texttt{test} set. 
$\dag$ denotes the \texttt{prototype} setting: The model is initialized from a FCOS3D~\cite{wang2021fcos3d} checkpoint trained on the nuScenes 3D detection dataset. 
$\ddag$ denotes the \texttt{improved} setting: A pretrained model from DD3D~\cite{park2021dd3d} is used, which includes external data from DDAD~\cite{packnet}. $*$ denotes backbone is pretrained on COCO~\cite{lin2014microsoft} and nuImage~\cite{nuscenes2019}. 
}
\label{table:state-of-the-art}
\renewcommand{\arraystretch}{1.0}
    \begin{tabular}{lC{0.0cm}C{1cm}C{1.2cm}C{0.5cm}C{0.15cm}C{0.75cm}C{0.75cm}C{0.75cm}C{0.75cm}C{0.75cm}}
    \hline

    \hline
    \textbf{Methods} &   & \textbf{Backbone} & \textbf{mAP}$\uparrow$  &\textbf{NDS}$\uparrow$  &   &\textbf{mATE}$\downarrow$   &\textbf{mASE}$\downarrow$   &\textbf{mAOE}$\downarrow$   &\textbf{mAVE}$\downarrow$   &\textbf{mAAE}$\downarrow$  \\
    \hline
    FCOS3D$^{\dag}$~\cite{wang2021fcos3d} & & R101 & 35.8 & 42.8 & & 69.0 & 24.9 & 45.2 & 143.4 & \textbf{12.4} \\
    PGD$^{\dag}$~\cite{wang2022probabilistic}& & R101 & 38.6 & 44.8 & & 62.6 & \textbf{24.5} & 45.1 & 150.9 & 12.7 \\
    Ego3RT$^{\dag}$~\cite{lu2022ego3rt} & & R101 & 38.9 & 44.3 & & \textbf{59.9} & 26.8 & 47.0 & 116.9 & 17.2 \\ 
    BEVFormer-S$^{\dag}$~\cite{li2022bevformer} & & R101 & 40.9 & 46.2 & & 65.0 & 26.1 & 43.9 & 92.5 & 14.7 \\
    \rowcolor[gray]{.9} 
    PolarFormer$^{\dag}$ & & R101 & \textbf{41.5} & \textbf{47.0} & & 65.7 & 26.3 & \textbf{40.5} & \textbf{91.1} & 13.9 \\
    
    \cmidrule(lr){1-11}
    BEVFormer$^{\dag}$~\cite{li2022bevformer} & & R101 &  44.5 & 53.5 & & 63.1 & 25.7 & 40.5 & \textbf{43.5} & 14.3 \\
    \rowcolor[gray]{.9} 
   
    \rowcolor[gray]{.9} 
    PolarFormer-T$^{\dag}$ & & R101 &  \textbf{45.7} & \textbf{54.3} & & \textbf{61.2} & \textbf{25.7} & \textbf{39.2} & 46.7 & \textbf{12.9} \\
    \cmidrule(lr){1-11}
   
    DETR3D$^{\ddag}$~\cite{wang2022detr3d} & & V2-99 & 41.2  & 47.9 & & 64.1 & 25.5 & 39.4 & 84.5 & 13.3\\
    
    M2BEV$^{*}$~\cite{xie2022m} & & X101 &  42.9 & 47.4 & & 58.3 & 25.4 & 37.6 & 10.53 & 19.0 \\
    Ego3RT$^{\ddag}$~\cite{lu2022ego3rt} & &V2-99 &  42.5 & 47.9 & & \textbf{54.9} & 26.4 & 43.3 & 101.4 & 14.5 \\ 
    BEVFormer-S$^{\ddag}$ & & V2-99 & 43.5  & 49.5 & & 58.9 & \textbf{25.4} & 40.2 & \textbf{84.2} & \textbf{13.1}\\
    \rowcolor[gray]{.9} 
    PolarFormer$^{\ddag}$ & & V2-99& \textbf{45.5}  & \textbf{50.3} & & 59.2 & 25.8 & \textbf{38.9} & 87.0 & 13.2 \\

    \cmidrule(lr){1-11}
    BEVFormer$^{\ddag}$~\cite{li2022bevformer} & & V2-99 & 48.1   & 56.9 & & 58.2 & \textbf{25.6} & 37.5 & \textbf{37.8} & \textbf{12.6}  \\
    
    \rowcolor[gray]{.9} 
    \rowcolor[gray]{.9} 
    PolarFormer-T$^{\ddag}$ & & V2-99 & \textbf{49.3} & \textbf{57.2} & &  \textbf{55.6} & \textbf{25.6} & \textbf{36.4} & 44.0 & 12.7  \\
    \hline
    
    \hline
    \end{tabular}
  \label{tab:nus-det-val}
\end{table*}%
 
\subsection{Polar BEV encoder at multiple scales}
We leverage multi-scale feature maps for handling object scale variance in the Polar coordinate. 
To that end,
the BEV Polar encoder performs information exchange among neighbouring pixels and across multi-scale feature maps. 
Formally, let $\{\mathbf{G}_u\}_{u=1}^U$
be the input multi-scale Polar feature maps and $\hat{x}_q \in [0, 1]^2$ be the normalized coordinates of the reference points for each qeury element $q$, we introduce a multi-scale deformable attention module \cite{zhu2020deformable} as:
\begin{equation}\label{eq:ms_deform_attn}
\begin{split}
    & \mathrm{MSDeformAttn}(\mathbf{z}_q,x_q, \{\mathbf{G}_u\}_{u=1}^U) = \\
    & \sum_{m=1}^M \mathbf{W}_m [\sum_{u=1}^{U}\sum_{k=1}^K A_{muqk}\mathbf{W'}_m\mathbf{G}_u(\zeta_u(\hat{x}_q)+\Delta x_{muqk})],
\end{split}
\end{equation}
where $m$ and $k$ are the index of the attention head and the sampling point. $\mathbf{z}_q$ is the query feature. $\Delta x_{muqk}$ and $A_{muqk}$ denote the sampling offset and the attention weight of the $k$th sampling point in $u$th feature level and $m$th attention head. The attention weight $A_{muqk}$ is normalized by $\sum_{u=1}^U\sum_{k=1}^K A_{muqk} = 1$. Sampling offsets $\Delta x_{muqk}$ are generated by applying MLP layers on query $q$. Function $\zeta_u$ generates the sampling offsets and rescales the normalized coordinate $\hat{x}_q$ to the $u$th feature scale. $\mathbf{W}_m$ and $\mathbf{W}'_m$ are learnable parameters. 
Serving as query, each pixel in the multi-scale feature maps exploits the information from both neighbouring pixels and pixels across scales.
This enables learning richer semantics across all feature scales.

\subsection{Polar BEV decoder at multiple scales}
The Polar decoder decodes the above multi-scale Polar features to make predictions in the Polar coordinate. 
We construct the Polar BEV decoder
with deformable attention \cite{zhu2020deformable}.
Specifically, we query $q$ in Eq. \eqref{eq:ms_deform_attn} as learnable parameters. 

Unlike 2D reference points in the encoder,
here the reference points are in 3D cylindrical coordinate, equal to Polar coordinate when projected to BEV.
The classification branch in each decoder layer outputs the confidence score vector $\mathbf{c}\in \mathbb{R}^{\mathcal{O}}$, where $\mathcal{O}$ is the number of categories. The key learning targets of regression branch are in polar coordinate instead of Cartesian coordiante, as illustrated in Figure \ref{fig:learning_targets}. 
For simplicity, superscript $^{(P)}$ is omitted. 
Reference points $(\rho,\phi, z)$ are iteratively refined in the decoder. 
With reference points, the regression branch regresses the offsets $d_\rho$, $d_\phi$ and $d_z$. 
The learning targets for orientation $\theta$ and velocity $v$ are relative to azimuths of objects and separated to orthogonal
components $\theta_{\phi}$, $\theta_{\rho}$, $v_{\phi}$ and $v_{\rho}$, defined by:
\begin{equation}
        \bar{\theta}_{ori} = \theta_{ori} - \phi,\quad \ 
        \theta_{\phi} = \sin{\bar{\theta}_{ori}},\quad \ 
        \theta_{\rho} = \cos{\bar{\theta}_{ori}},
\end{equation}
and
\begin{equation}
        \bar{\theta}_{v} = \theta_{v} - \phi, \quad \ 
        v_{\phi} = v_{abs}\sin{\bar{\theta}_{v}}, \quad \ 
        v_{\rho} = v_{abs}\cos{\bar{\theta}_{v}}.
\end{equation}
Here, $\theta_{ori}$ is the yaw angle of the bounding box. $v_{abs}$ and $\theta_v$ are the absolute value and angle of velocity. We regress the object size $l$, $w$ and $h$ as $\log l$, $\log w$ and $\log h$. 
We adopt Focal loss \cite{lin2017focal} and L1 loss for classification and regression respectively.

\newcommand{\lack}[1]{\textcolor{cyan}{#1}}
\newcommand{\bug}[1]{\textcolor{brown}{#1}}
\newcommand{\es}[1]{\textcolor{magenta}{#1}}

\section{Experiments}
\label{sec:exp}
\begin{table*}[t]
    \footnotesize
    \caption{3D object detection results in different coordinate systems and ablations for the model architecture. 
    PC denotes feature in Polar and prediction in Cartesian.
    }
    \vspace{-10pt}
    \begin{subtable}[h]{0.65\textwidth}
        \centering
        \setlength{\tabcolsep}{4.8pt}
        \renewcommand{\arraystretch}{0.97}{
        \begin{tabular}{l||ll||ll}
        \hline
        
        \hline
        \rowcolor{white}
        \textbf{Method} & \textbf{Feature} & \textbf{Prediction} & \textbf{mAP}$\uparrow$ & \textbf{NDS}$\uparrow$ \\
        \hline
        Centerpoint~\cite{yin2021center} & Cartesian & Cartesian & 37.8 & 45.4  \\
        Centerpoint*~\cite{yin2021center} & Cartesian & Cartesian & 38.5 & 45.6  \\
        PolarFormer-CC& Cartesian & Cartesian & 38.1 & 45.5 \\
        PolarFormer-PC& Polar & Cartesian & 38.5 & 45.0 \\
        \rowcolor[gray]{.9} 
        PolarFormer & Polar& Polar& \textbf{39.6} & \textbf{45.8} \\
        \hline
        
        \hline
        \end{tabular}
        }
        \caption{Ablation study on coordinate system and detection head.}
        \label{table:coordinate}
    \end{subtable}
    \hfill
    \begin{subtable}[h]{0.3\textwidth}
    \centering
    \setlength{\tabcolsep}{4.3pt}
    \renewcommand{\arraystretch}{1.45}{
    \begin{tabular}{l||ll}
    \hline
    
    \hline
    \textbf{Position encoding} & \textbf{mAP}$\uparrow$ & \textbf{NDS}$\uparrow$ \\
    \hline
    2D learnable PE &   38.8 & 45.1  \\
    3D PE &  38.5 & 44.9 \\
    \rowcolor[gray]{.9} 
    Fixed Sine PE &  \textbf{39.6} & \textbf{45.8} \\
    \hline
    
    \hline
    \end{tabular}
    }
    \vspace{6pt}
    \caption{Ablation on positional encoding (PE).}
    \label{table:pe}
    \end{subtable}
    \vfill
        \begin{subtable}[h]{0.65\textwidth}
        \centering
        \setlength{\tabcolsep}{15pt}
        \renewcommand{\arraystretch}{0.9}{
        \begin{tabular}{l||c||lll}
        \hline
        
        \hline
        \textbf{Methods} & \textbf{Multi-scale} &\textbf{mAP}$\uparrow$ & \textbf{NDS}$\uparrow$ & \textbf{mAOE}$\downarrow$ \\
        \hline
        PolarFormer-CC-s & \XSolidBrush & 38.1 & 44.9 & 40.0 \\
        PolarFormer-PC-s & \XSolidBrush & 38.8 & 45.0 & 37.6 \\
        \rowcolor[gray]{.9} 
        PolarFormer-s & \XSolidBrush & \textbf{39.1} & \textbf{45.0} & \textbf{37.3} \\
        \cmidrule(lr){1-5}        
        PolarFormer-CC &  \Checkmark & 38.1 & 45.0 & 37.5 \\
        PolarFormer-PC & \Checkmark & 38.5 & 45.5 & 40.8 \\
        \rowcolor[gray]{.9} 
        PolarFormer & \Checkmark & \textbf{39.6} & \textbf{45.8} & \textbf{37.5} \\
        \hline
        
        \hline
        \end{tabular}
        }
        \caption{Effectiveness of multi-scale polar representation.}
        \label{table:multi-scale}
    \end{subtable}
    \hfill
    \begin{subtable}[h]{0.3\textwidth}
        \centering
        \setlength{\tabcolsep}{6.15pt}
        \renewcommand{\arraystretch}{1.12}{
        \begin{tabular}{lll||ll}
        \hline
        
        \hline
        $\bm{\mathcal{N}_1}$ & $\bm{\mathcal{R}_1}$ & &  \textbf{mAP}$\uparrow$ & \textbf{NDS}$\uparrow$\\
        \hline
        240  & 64 & & 38.8 & 45.0 \\
        \rowcolor[gray]{.9} 
        256  & 64 & & \textbf{39.6} & \textbf{45.8} \\
        272  & 64 & & 38.8 & 45.0  \\
        \cmidrule(lr){1-5}
        256  & 56 & & 38.7 & 45.4 \\
        256  & 72 & & 38.5 & 45.4 \\
        256  & 80 & & 38.7 & 45.6 \\
        \hline
        
        \hline
        \end{tabular}
        }
        \caption{Ablation on polar resolution.}
        \label{table:resolution}
    \end{subtable}
\vspace{-10pt}
\end{table*}

\paragraph{Dataset}
We evaluate the PolarFormer on the nuScenes dataset~\cite{nuscenes2019}.
It provides images with a resolution of $1600 \times 900$ from 6 surrounding cameras (Figure~\ref{fig:teaser}). 
The total of 1000 scenes, where each sequence is roughly 20 seconds long and annotated every 0.5 second, is split officially into \texttt{train/val/test} set with 700/150/150 scenes. 
\paragraph{Implementation details}
We implement our approach based on the codebase \texttt{mmdetection3d}~\cite{mmdet3d2020}.
Following DETR3D~\cite{wang2022detr3d} and FCOS3D~\cite{wang2021fcos3d}, a ResNet-101~\cite{he2016deep}, with 3rd and 4th stages equipped with deformable convolutions is adopted as the backbone architecture. 
The number of cross-plane encoder layer is set to 3 for each feature scale.
The resolution of radius and azimuth for our multi-scale Polar BEV maps are (64,~256), (32,~128), (16,~64) respectively.
We use 6 Polar BEV encoder and 6 decoder layers.
Following DETR3D~\cite{wang2022detr3d}, our backbone is initialized from a checkpoint of FCOS3D~\cite{wang2021fcos3d} trained on nuScenes 3D detection task, while the rest is initialized randomly. 
We use above setting for \texttt{prototype} verification.
To fully leverage the sequence data, we further conduct temporal fusion between the current frame and one history sweep in the BEV space. 
Following BEVDet4D~\cite{huang2022bevdet4d}, we simply concatenate two temporally adjacent multi-scale Polar BEV maps along the feature dimension and feed to the BEV Polar encoder.
We randomly sample a history sweep from [3T; 27T] during training,
and sample the frame at 15T for inference. 
T ($ \approx 0.083s$) refers to the time interval between two sweep frames.
We term our temporal version as \texttt{PolarFormer-T}.

\paragraph{Training }
Following DETR3D~\cite{wang2022detr3d} we train our models for 24 epochs with the AdamW optimizer and cosine annealing learning rate scheduler on 8 NVIDIA V100 GPUs.
The initial learning rate is $2 \times 10^{-4}$, and the weight decay is set to $0.075$. 
Total batch size is set to 48 across six cameras.
Synchronized batch normalization is adopted.
All experiments use the original input resolution.
Note our image variant uses the same amount of training data as the monocular pipelines~\cite{wang2021fcos3d} and the multi-camera counterparts~\cite{wang2022detr3d,li2022bevformer}.
Multi-camera and monocular paradigms share the same evaluation metrics.

\paragraph{Inference } 
We evaluate our model on nuScenes validation set and test server.
We do not adopt model-agnostic trick such as model ensemble and test time augmentation.

\subsection{Comparison with the state of the art}

We compare our method with the state of the art on both \texttt{test} and \texttt{val} sets of nuScenes. 
In addition to the 
{\bf (i)} 
\texttt{prototype} setting mentioned in implementation details,
we also evaluate our model in the
{\bf (ii)}
\texttt{improved} setting, with VoVNet (V2-99)~\cite{lee2019energy} as backbone architecture with a pretrained checkpoint from DD3D~\cite{park2021dd3d} (fine-tuned on extra DDAD15M~\cite{packnet} dataset) to boost performance.

Table~\ref{table:state-of-the-art} compares the results on nuScenes \texttt{test} set.
We observe that our \texttt{PolarFormer} achieves the best performance under both the 
{\bf (i)}
\texttt{prototype} and
{\bf (ii)}
\texttt{improved} setting
{
in terms of \texttt{mAP} and \texttt{NDS} metrics, indicating the superiority of learning representation in the Polar coordinate. 
}
With temporal information \texttt{PolarFormer-T} can further boot performance substantially. 
Additional experiments results on \texttt{val} set t and qualitative
results are shown in supplementary materials.

\subsection{Ablation studies}
\begin{figure*}[t]

\begin{minipage}[b]{.27\linewidth}
    \centering
    \subfloat[][]{\label{cartesian_det}\includegraphics[ width=0.985\linewidth]{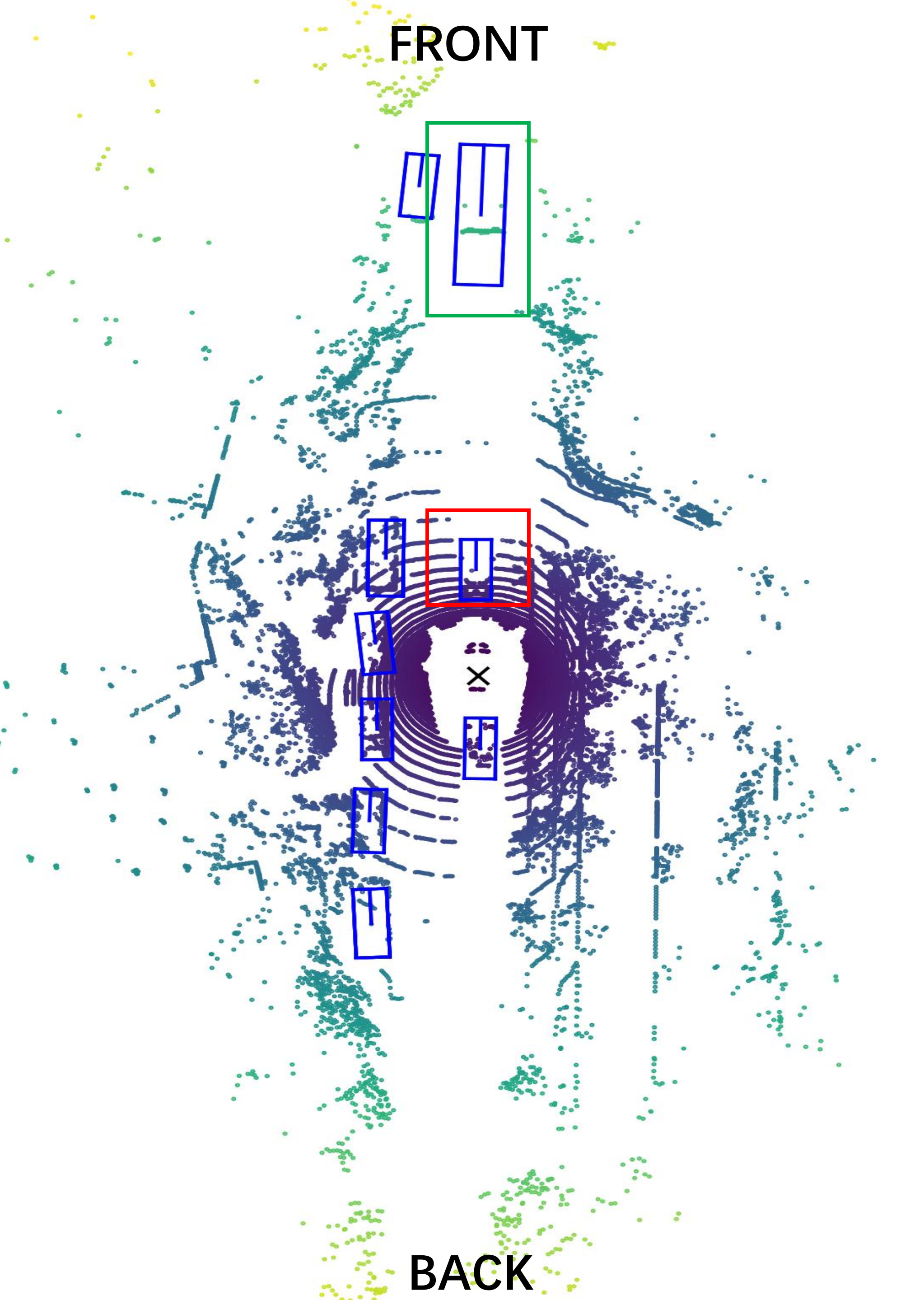}}
\end{minipage}
\begin{minipage}[b]{.69\linewidth}
    \centering
    \subfloat[][]{\label{polar_det}\includegraphics[width=0.9\linewidth]{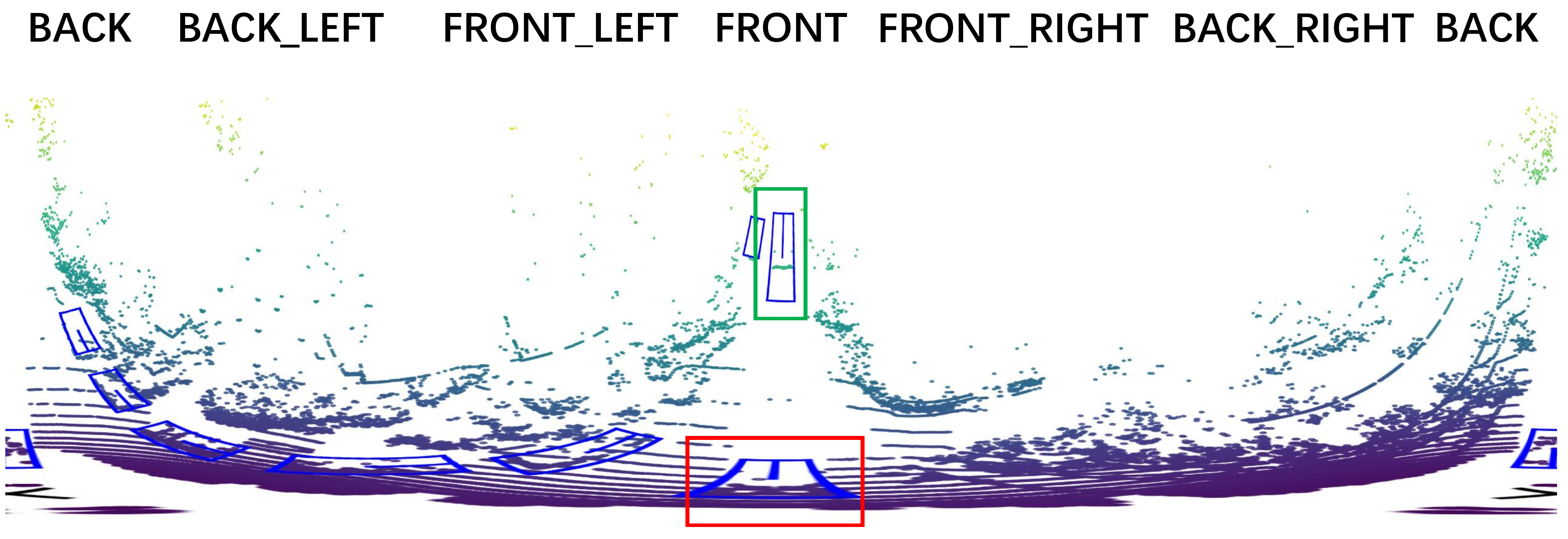}}
    
    \subfloat[][]{\label{diff_dist}
    \small
    \begin{subtable}[h]{\textwidth}
        \centering
        \setlength{\tabcolsep}{4pt}
        \renewcommand{\arraystretch}{1.2}{
        \begin{tabular}{ll||ccc}
        \hline
        
        \hline
        \rowcolor{white}
        \textbf{Method} & \textbf{Coordinate} & \textbf{Near} & \textbf{Medium} & \textbf{Far} \\
        \hline
        DETR3D~\cite{wang2022detr3d} & Cartesian & 49.9/49.3 & 28.6/39.2 & 8.1/21.2 \\
        BEVFormer-S~\cite{li2022bevformer} & Cartesian & 54.2/53.5 & 31.7/41.3 & 9.5/21.9  \\
        PolarFormer-CC & Cartesian & 54.7/55.5 & 32.3/42.2 & 9.4/22.2\\
        \rowcolor[gray]{.9} 
        PolarFormer & Polar & \textbf{57.8/55.8} & \textbf{33.6/42.9} & \textbf{9.6/22.3}  \\
        \hline
        
        \hline
        \end{tabular}
        }
        \label{table:coordinate}
    \end{subtable}
    
    }
\end{minipage} 
\caption{
3D object detection in (a) Cartesian BEV {\em vs.} (b) Polar BEV, and (c) Performance comparison (mAP/NDS) at three distances (Near/Medium/Far). 
Red and green boxes show the same objects in different coordinates.
}
\label{fig:polarcardi}
\end{figure*}


\begin{figure}[ht]
\centering
 \includegraphics[height=3.0cm, width=0.95\linewidth]{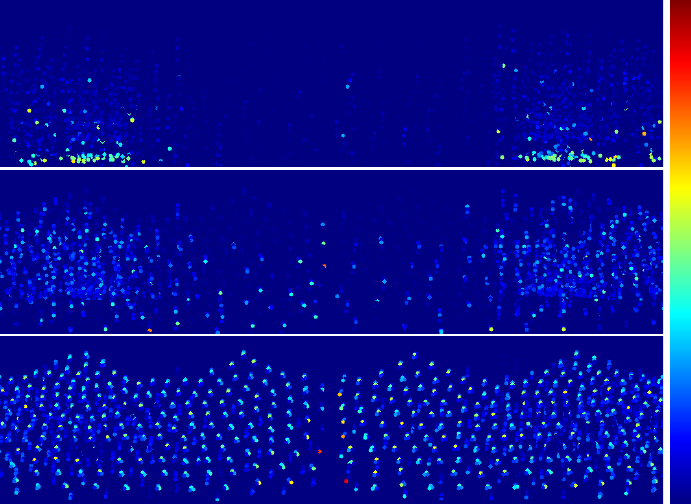}
 \caption{Multi-scale Polar BEV attention.}
 \label{fig:query_attn}
\end{figure}
We conduct a series of ablation studies on nuScenes \texttt{val} set to 
validate the design of PolarFormer.
Each proposed component and important hyperparameters are examined thoroughly.

\paragraph{Polar \textit{v.s.} Cartesian }
Table~\ref{table:coordinate} ablates the coordinate system.
We make several observations:
\textbf{(I)}
Learning the representation and making the prediction both on Cartesian, Centerpoint~\cite{yin2021center} gives a strong baseline with 0.378 mAP and 0.454 NDS;
After applying circle NMS, Centerpoint* can further improve;
\textbf{(II)}
Our PolarFormer-CC (with Cartesian feature and prediction) outperforms the Centerpoint equipped with specially designed CBGS head~\cite{yin2021center};
\textbf{(III)}
When Polar BEV map is used to feed into a Cartesian decoder head, \textit{on-par} performance is achieved with the post-processing counterpart;
\textbf{(IV)}
PolarFormer, our full model that predicts all 10 categories with one head and without any post-processing procedure, exceeds highly optimized Centerpoint by 1.1\% in mAP and 0.2\% in NDS.
This suggests the significance of \textit{Polar} 
in both representation learning and exploitation (\ie, decoding).

\paragraph{Visualizations }
With the quantitative evaluation in Figure~\ref{fig:polarcardi}, \textbf{(I)}
it is evident that our model in Polar coordinate yields better results than Cartesian consistently.
Specifically, Polar outperforms Cartesian by mAP 3.1\% and NDS 0.3\% in nearby area, mAP 1.3\% and NDS 0.7\% in medium area.
\textbf{(II)}
As shown in Figure~\ref{cartesian_det} and Figure~\ref{polar_det}, compared to the Polar map, Cartesian usually downsamples the nearby area (red) with information loss, while upsamples the distant area (green) without actual information added.
This would explain the inferiority of Cartesian.
\textbf{(III)}
Figure~\ref{fig:query_attn} shows the attention map of the decoder query in multi-scale Polar BEV features.
For better viewing, we resize the multi-scale features into the same resolution.
The bottom/top corresponds to the largest/smallest scale features.
The \textit{y} axis represents the radius of the Polar map.
It is shown that larger objects represented in the small map (top) are close to the ego car (small radius) whilst small objects in the large map (bottom) distribute through the distant area.
This is highly consistent with the geometry structure of raw images
(Figure~\ref{fig:teaser}), which has shown to be a more effective coordinate 
for 3D object detection as above.

\paragraph{Architecture }
\textbf{(I)}
We first evaluate three designs of positional embedding (PE):
2D learnable PE, fixed Sine PE, and 3D PE (generated based on a set of 3D points for each Polar ray position).
\textbf{(II)}
As our cross-plane encoder transforms different levels of feature from FPN into Polar rays independently, we can fuse the multi-level features into a single BEV or naturally shape multiple BEVs with different \textit{or} same resolutions;
Table~\ref{table:multi-scale} clearly shows that a model with multi-scale Polar BEVs outperforms the single-scale counterpart under either coordinate.
In contrast, little performance gain is achieved from multi-scale features in Cartesian.
This suggests that object scale variation is a {\em unique} challenge with Polar, but absent with Cartesian.
Our design consideration is thus verified.
\textbf{(III)} 
We study the resolution of polar map by adjusting the \textit{azimuth} $\bm{\mathcal{N}_1}$ and \textit{radius} $\bm{\mathcal{R}_1}$  (the number of Polar query in the cross-plane encoder);
Table \ref{table:resolution} shows that the \textit{angle} with 256 and \textit{radius} with 64 gives the best performance.
\section{Conclusions}

We have proposed the Polar Transformer (PolarFormer) for 3D object detection in multi-camera 2D images from the ego car's perspective.
With a rasterized BEV Polar representation geometrically aligned to visual observation, 
PolarFormer overcomes irregular Polar grids by a cross-attention based decoder.
Further, a multi-scale representation learning strategy is designed 
for tackling the intrinsic object scale variation challenge.
Extensive experiments on the nuScenes dataset validate 
the superiority of our PolarFormer over previous alternatives on 3D object detection.

\vspace{2mm}
\noindent \textbf{Acknowledgment} This work was supported by the National Key R$\&$D Program of China (Grant No. 2018AAA0102803, 2018AAA0102802, 2018AAA0102800), 
the Natural Science Foundation of China (Grant No. 6210020439, U22B2056, 61972394, 62036011, 62192782, 61721004, 62102417), 
Lingang Laboratory (Grant No. LG-QS-202202-07),
Natural Science Foundation of Shanghai (Grant No. 22ZR1407500)
Beijing Natural Science Foundation (Grant No. L223003, JQ22014), 
the Major Projects of Guangdong Education Department for Foundation Research and Applied Research (Grant No. 2017KZDXM081, 2018KZDXM066), 
Guangdong Provincial University Innovation Team Project (Grant No. 2020KCXTD045). 
Jin Gao was also supported in part by the Youth Innovation Promotion Association, CAS.
\bibliography{main}

\appendix

\section{Appendix}

\subsection{Experiments on the val set of nuScenes}
\begin{table*}[t]
\footnotesize
  \centering
    \caption{
    State-of-the-art comparison on nuScenes \texttt{val} set. 
    $\dag$ denotes the \texttt{prototype} setting: The model is initialized from a FCOS3D~\cite{wang2021fcos3d} checkpoint trained on the nuScenes 3D detection dataset. 
    $\ddag$ denotes \texttt{improved} setting: A pretrained model from DD3D~\cite{park2021dd3d} is used, which includes extra data from DDAD~\cite{packnet}.$*$ denotes backbone is pretrained on COCO~\cite{lin2014microsoft} and nuImage~\cite{nuscenes2019}.}

\label{table:nusc_val}
\renewcommand{\arraystretch}{1.2}
    \begin{tabular}{lC{0.0cm}C{1cm}C{1.2cm}C{0.5cm}C{0.15cm}C{0.75cm}C{0.75cm}C{0.75cm}C{0.75cm}C{0.75cm}}
    \hline

    \hline
    \textbf{Methods} &   & \textbf{Backbone} & \textbf{mAP}$\uparrow$  &\textbf{NDS}$\uparrow$  &   &\textbf{mATE}$\downarrow$   &\textbf{mASE}$\downarrow$   &\textbf{mAOE}$\downarrow$   &\textbf{mAVE}$\downarrow$   &\textbf{mAAE}$\downarrow$  \\
    \hline
    FCOS3D$^{\dag}$~\cite{wang2021fcos3d} & & R101 & 32.1 &  39.5 & & 75.4 & \textbf{26.0} & 48.6 & 133.1 & \textbf{15.8} \\
    DETR3D$^{\dag}$~\cite{wang2022detr3d} & & R101 & 34.7 & 42.2 & & 76.5 & 26.7 & 39.2 & 87.6 & 21.1\\
    PGD$^{\dag}$~\cite{wang2022probabilistic} & & R101& 35.8  & 42.5 & & 66.7 & 26.4 & 43.5 & 127.6 & 17.7\\
    Ego3RT$^{\dag}$~\cite{lu2022ego3rt} & & R101 &  37.5 & 45.0 & & \textbf{65.7} & 26.8 & 39.1 & 85.0 & 20.6 \\ 
    BEVFormer-S$^{\dag}$~\cite{li2022bevformer} & & R101 & 37.5 & 44.8 & & 72.5 & 27.2 & 39.1 & \textbf{80.2} & 20.0\\
    \rowcolor[gray]{.9} 
    PolarFormer$^{\dag}$ & & R101 & \textbf{39.6} & \textbf{45.8} & & 70.0 & 26.9 & \textbf{37.5} & 83.9 & 24.5\\
    \cmidrule(lr){1-11}
    Ego3RT$^{\ddag}$~\cite{lu2022ego3rt} & & V2-99 &  47.8 & 53.4 & & \textbf{58.2} & 27.2 & 31.6 & 68.3 & 20.2 \\ 
    M2BEV$^{*}$~\cite{xie2022m} & & X101 &  41.7 & 47.0  & & 64.7 & 27.5 & 37.7 & 83.4 & 24.5 \\
    \rowcolor[gray]{.9} 
    PolarFormer$^{\ddag}$ & & V2-99 & \textbf{50.0}  & \textbf{56.2} & & 58.3 & \textbf{26.2} & \textbf{24.7} & \textbf{60.1} & \textbf{19.3} \\
    \cmidrule(lr){1-11}
     BEVFormer$^{\dag}$~\cite{li2022bevformer} & & R101 & 41.6 & 51.7 & & 67.3 & 27.4 & 37.2 & \textbf{39.4} & \textbf{19.8} \\
 
     \rowcolor[gray]{.9} 
     PolarFormer-T$^{\dag}$ & & R101 & \textbf{43.2} & \textbf{52.8} & & \textbf{64.8} & \textbf{27.0} & \textbf{34.8} & 40.9 & 20.1 \\
     \rowcolor[gray]{.9} 
    \hline
    
    \hline
    \end{tabular}
  \label{tab:nus-det-val}
\end{table*}%
Table~\ref{table:nusc_val} shows that our method achieves leading performance on the \texttt{val} set for both \texttt{mAP} and \texttt{NDS} metrics.
Under the \texttt{improved} setting, PolarFormer shines on all metrics except \texttt{mASE} and \texttt{mAAE}.
Again, \texttt{PolarFormer-T} outperforms the alternative BEVFormer by a clear margin,
indicating the superiority of learning representation in the Polar coordinate. 

\begin{table}[]
    \footnotesize
    \caption{
    Ablation study on the coordinate of 
    {\em object location} prediction and loss calculation.
    }
    \label{tab:coord_target_loss}
    \centering
    \setlength{\tabcolsep}{12pt}
    \renewcommand{\arraystretch}{1.1}{
    \begin{tabular}{ll||ll}
    \hline
    
    \hline
    \cellcolor{white}\textbf{Prediction} & \textbf{Loss} & \textbf{mAP}$\uparrow$ & \textbf{NDS}$\uparrow$ \\
    \hline
    Cartesian & Cartesian & 38.0 & 44.9  \\
    Polar & Polar & 31.5 & 38.2 \\
    \rowcolor[gray]{.9} 
    Polar & Cartesian &  \textbf{39.6} & \textbf{45.8} \\
    \hline
    
    \hline
    \end{tabular}
    }
\end{table}

\subsection{The coordinate choice for object prediction and loss optimization}

Until now, we have shown that making the predictions in the Polar coordinate is crucial and superior over in the Cartesian counterpart.
We conjugate that 
this is because with object locations projected under the Polar coordinate, a better distribution of reference points can be learned due to taking a more consistent perspective \wrt{} the multi-camera image observation.
Interestingly, we note a discrepancy in the coordinate choice 
between {\em object location} prediction and loss optimization.
In particular, we find that when optimizing the object localization loss in the Polar coordinate, the loss converges very slowly and a significant performance degradation is also observed (Table \ref{tab:coord_target_loss}). 
A plausible obstacle is the numerical discontinuity between 0 and $2\pi$ in azimuth, which however is continuous in the physical world. 

\subsection{Visualization}
\begin{figure*}[h]
     \centering
     \begin{subfigure}[b]{\textwidth}
         \centering
         \makebox[\textwidth][c]{\includegraphics[width=1.\textwidth]{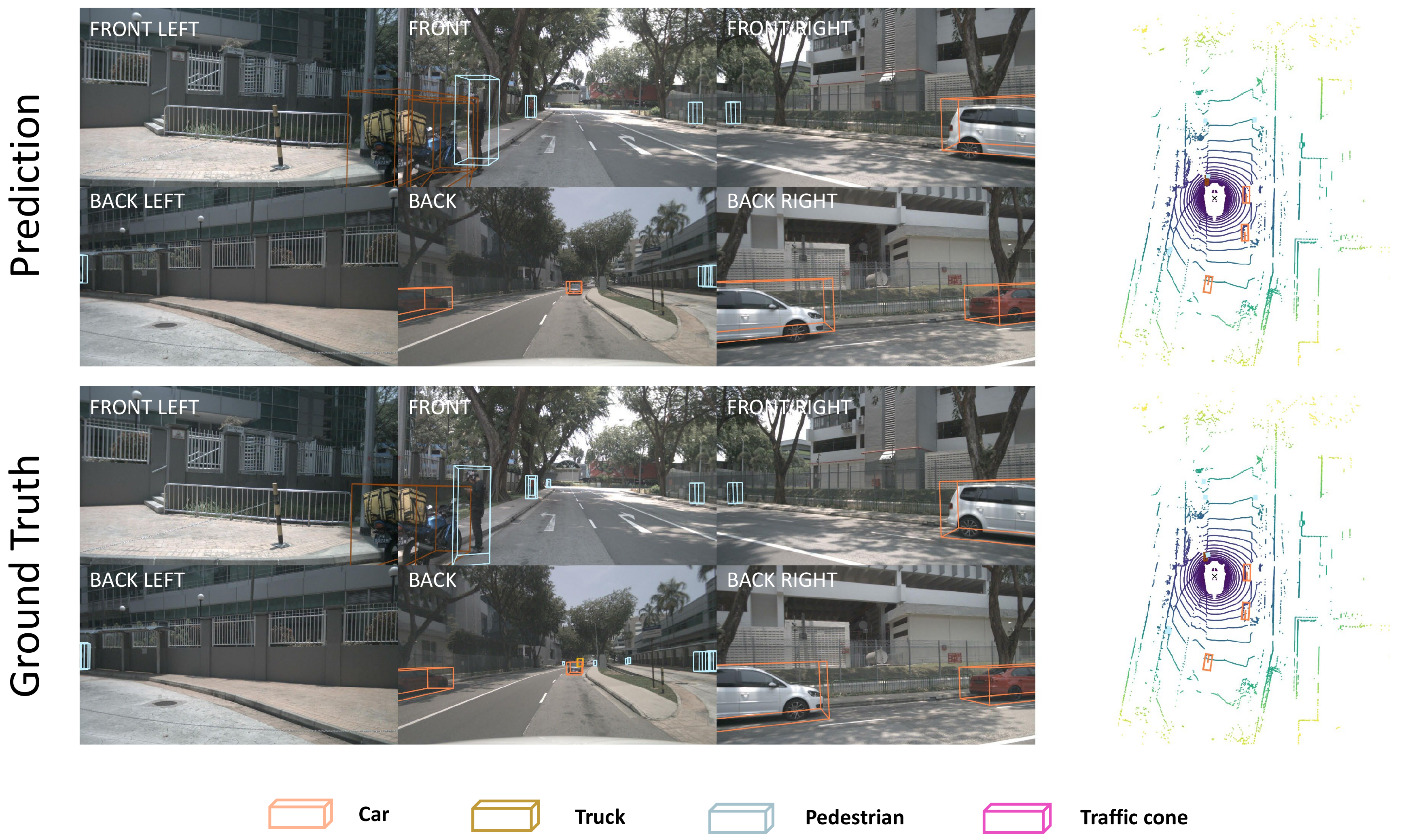}}
         \label{fig:1}
     \end{subfigure}
     \vfill
     \begin{subfigure}[b]{\textwidth}
         \centering
         \makebox[\textwidth][c]{\includegraphics[width=1.\textwidth]{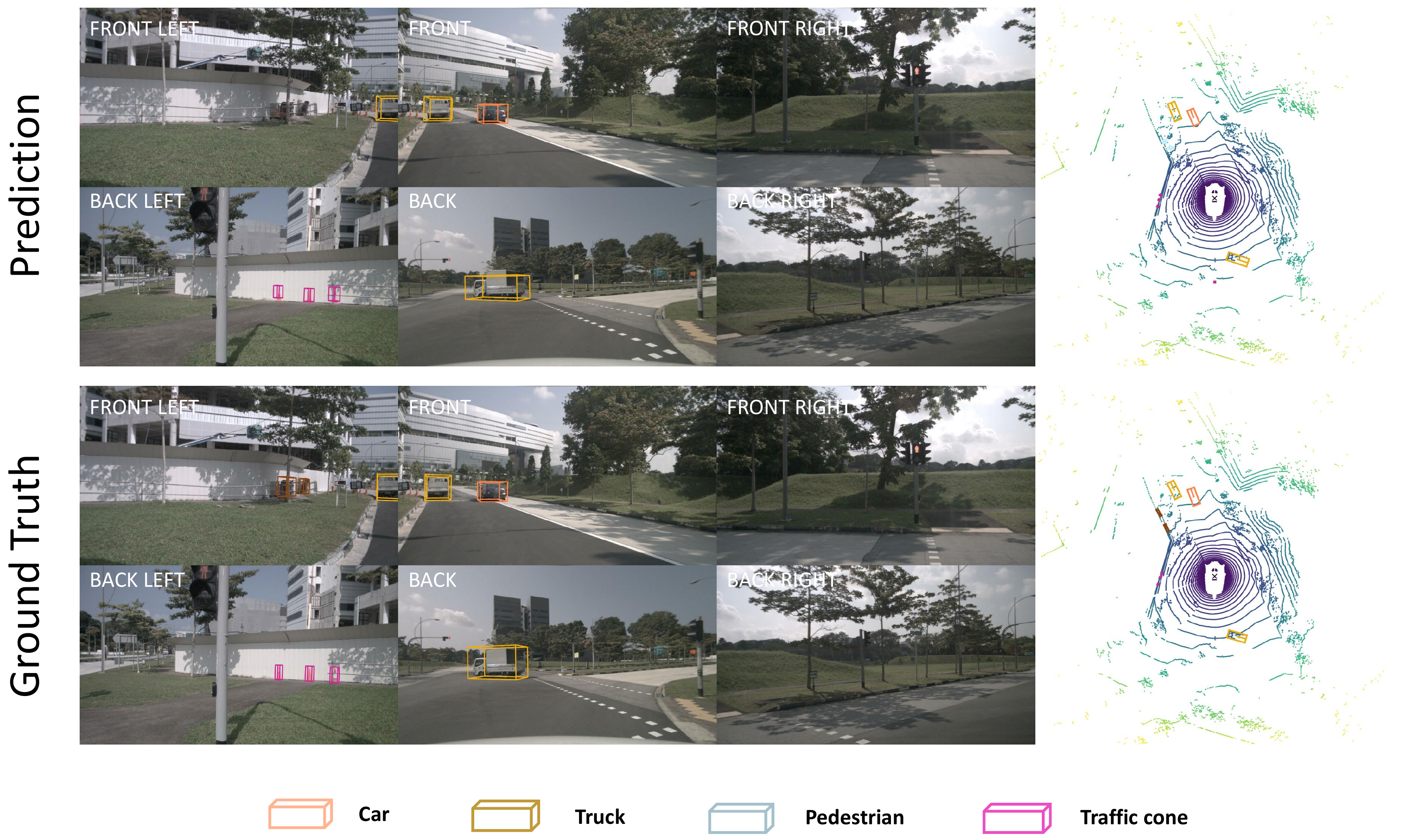}}
         \label{fig:2}
     \end{subfigure}
     \caption{Visualizations (Part I): Given multi-camera images, our method dedicates to perform 3D object detection and BEV semantic segmentation.}
     \label{fig:supp_vis_1}
\end{figure*}

\begin{figure*}[htbp]
     \centering
     \begin{subfigure}[b]{\textwidth}
         \centering
         \makebox[\textwidth][c]{\includegraphics[width=1.\textwidth]{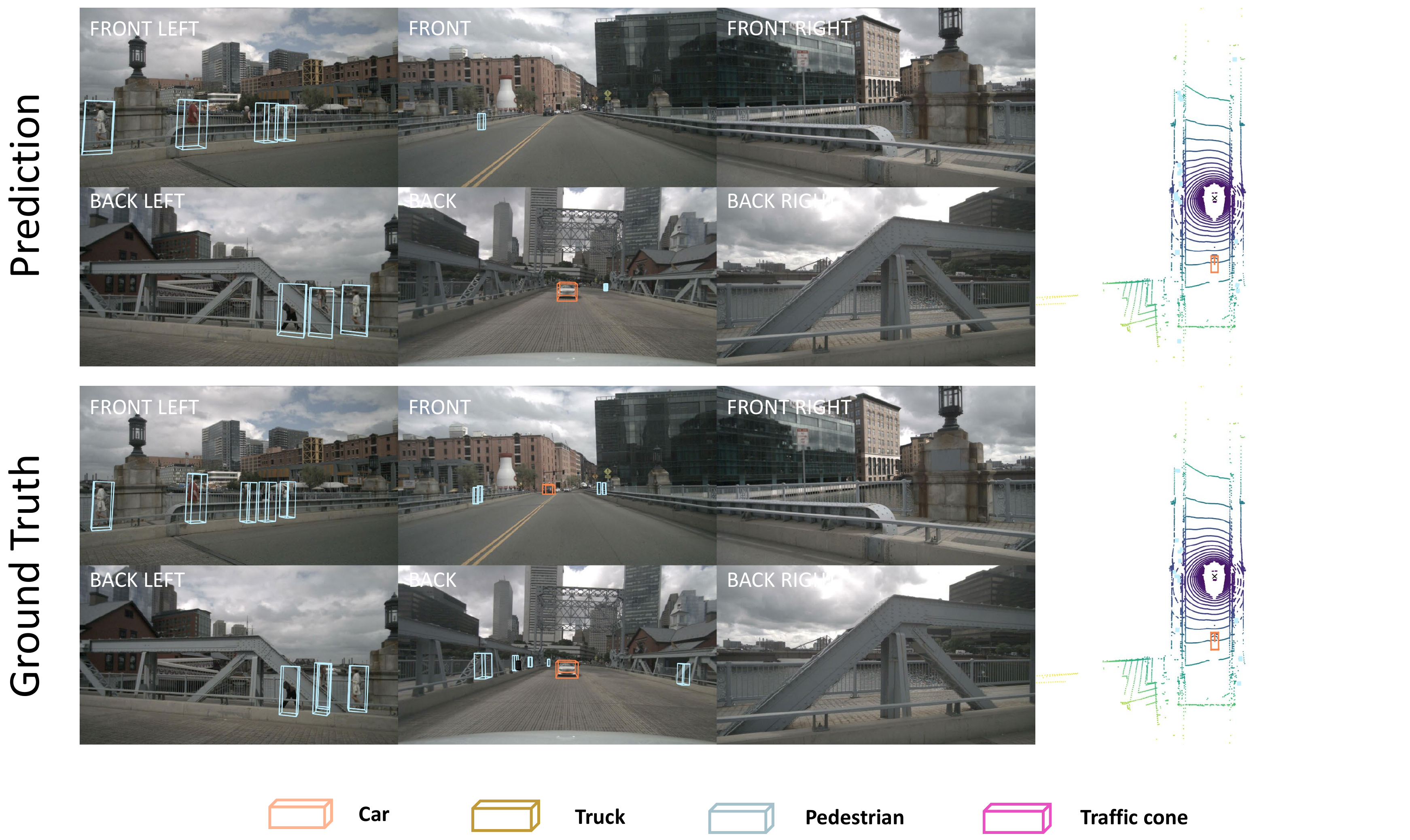}}
         \label{fig:3}
     \end{subfigure}
     \vfill
     \begin{subfigure}[b]{\textwidth}
         \centering
         \makebox[\textwidth][c]{\includegraphics[width=1.\textwidth]{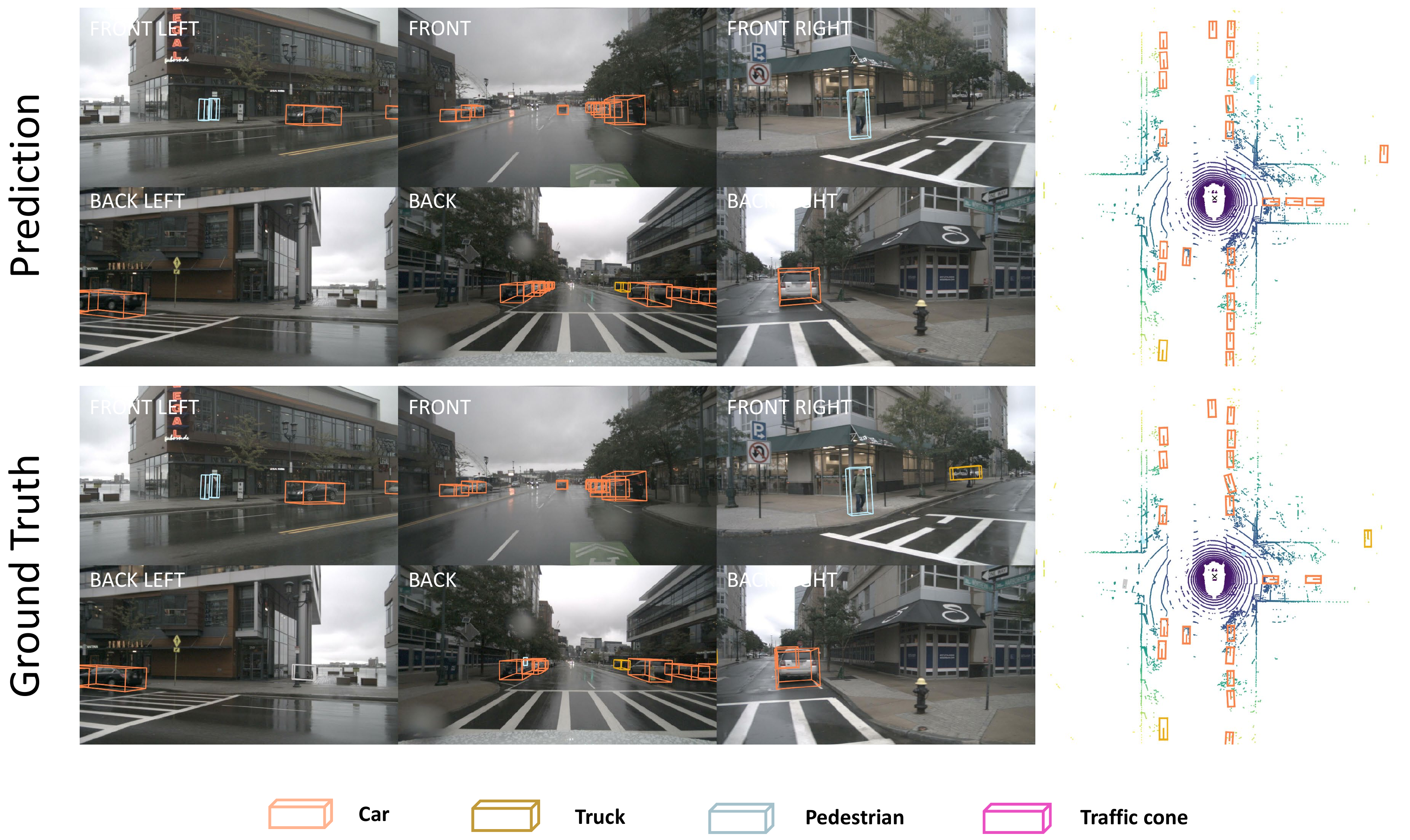}}
         \label{fig:4}
     \end{subfigure}
      \caption{Visualizations (Part II): Given multi-camera images, our method dedicates to perform 3D object detection and BEV semantic segmentation.}
     \label{fig:supp_vis_2}
\end{figure*}

\begin{figure*}[htbp]
     \centering
     \begin{subfigure}[b]{\textwidth}
         \centering
         \makebox[\textwidth][c]{\includegraphics[width=1.\textwidth]{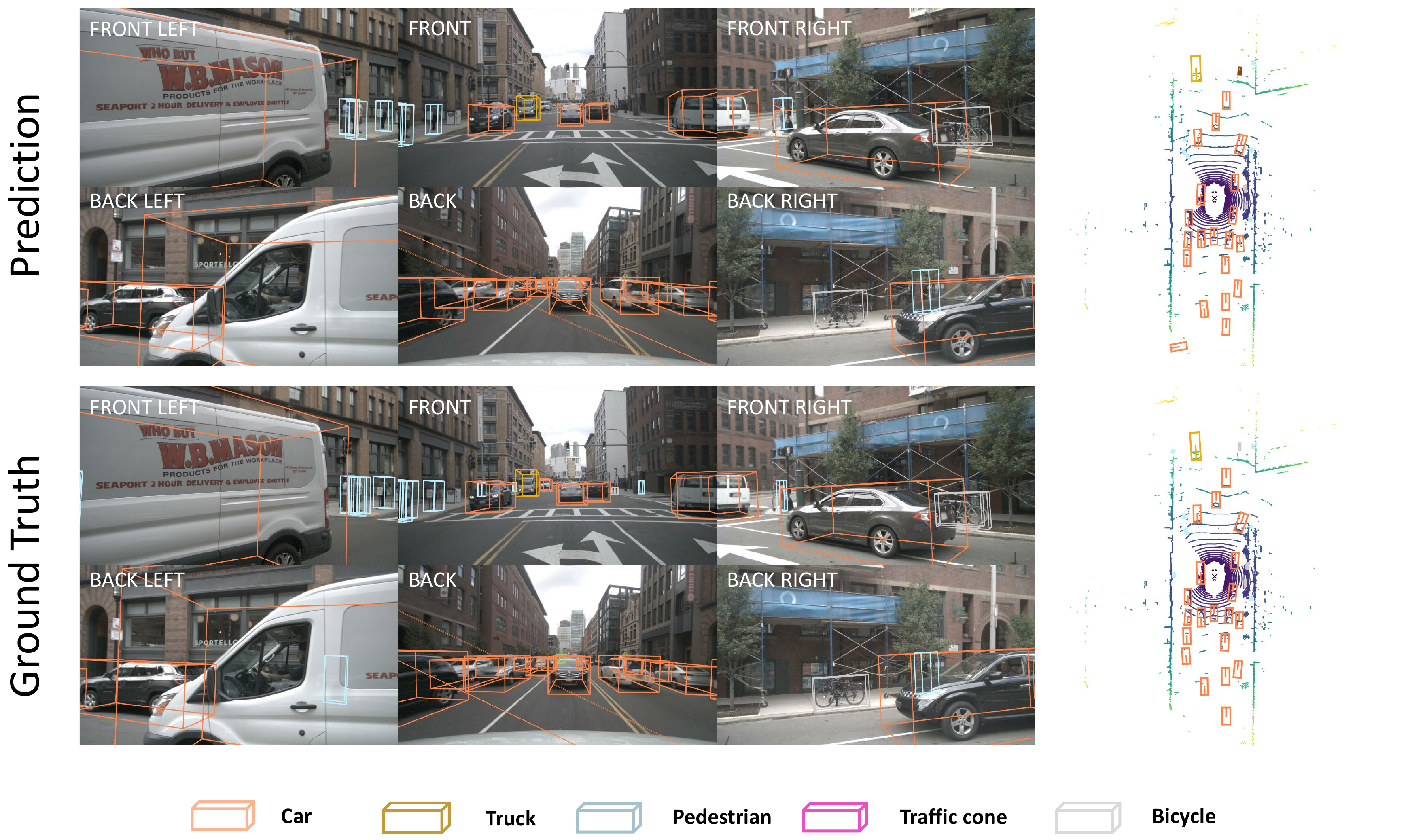}}
         \label{fig:4}
     \end{subfigure}
      \caption{Visualizations (Part III): Given multi-camera images, our method dedicates to perform 3D object detection and BEV semantic segmentation.}
     \label{fig:supp_vis_3}
\end{figure*}

We visualize our 3D object detection and BEV semantic segmentation results in Figure~\ref{fig:supp_vis_1}, Figure~\ref{fig:supp_vis_2} and Figure~\ref{fig:supp_vis_3}.

\subsection{Limitations and potential societal impact}
\paragraph{Limitations} Since existing BEV based 3D object detection methods 
consider usually the Cartesian coordinate,
there is a lacking of
elaborately designed off-the-shelf LiDAR detection heads and corresponding tricks suitable for the Polar coordinate based methods as we propose here.
That means there is some room for performance gain 
in head design, which will be one of the future works.

\paragraph{Societal impact} 
Our method can be used as the perception module for autonomous driving. However, our system is not perfect yet and hence not fully trustworthy in real-world deployment. Also, the current system is not exhaustively evaluated and tested due to limited resources as existing alternative works.
Autonomous driving is still a largely immature field with many ongoing and unsolved matters including those complex legitimate issues, and many of those may be concerned with this work too.

\subsection{Scale variance of objects on Polar BEV feature map}
Although Polar BEV feature map could preserve rich features in non-far regions when compared with Cartesian BEV feature map, it has to overcome scale variance problem, a challenge which is not presented in Cartesian BEV feature map. The object scale variance lies in that the size of an object in the Polar BEV feature map could reduce increasingly when it moves away from the ego-vehicle. Here we give an example of objects moving far away from the ego-vehicle in radial direction and provide simple mathematical proof. 

\begin{figure*}
     \centering
     \begin{subfigure}[b]{0.425\textwidth}
         \centering
         \includegraphics[width=\textwidth]{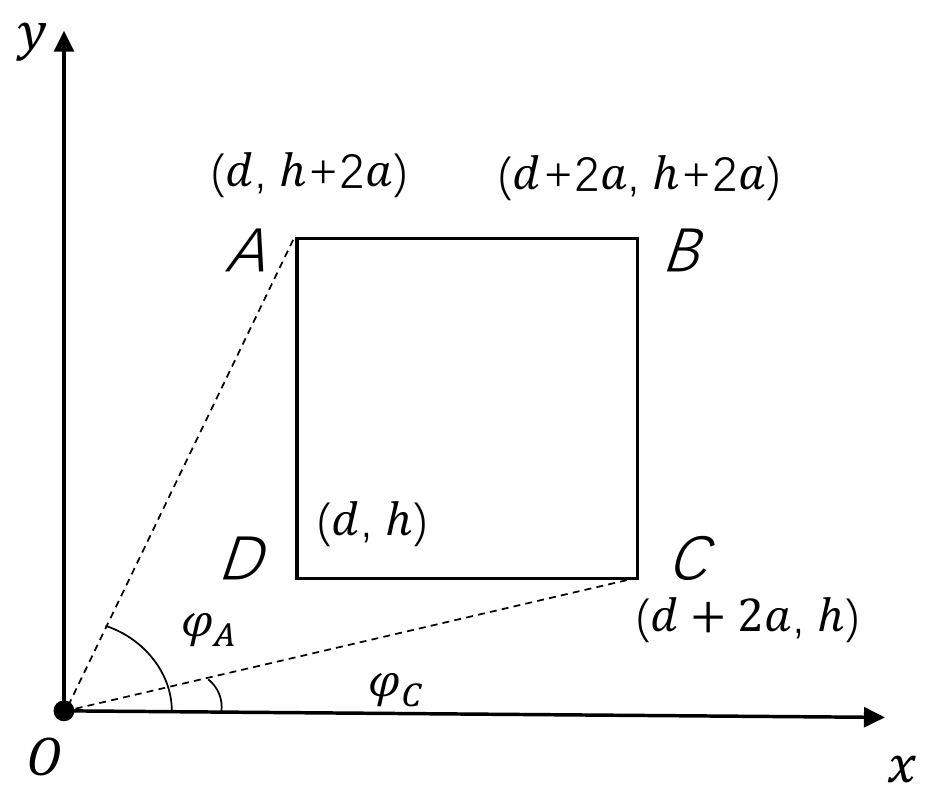}
         \caption{Cartesian BEV feature map}
         \label{fig:supp_cartesian_coord}
     \end{subfigure}
     \hfill
     \begin{subfigure}[b]{0.56\textwidth}
         \centering
         \includegraphics[width=\textwidth, trim={0 0.11cm 0 0}]{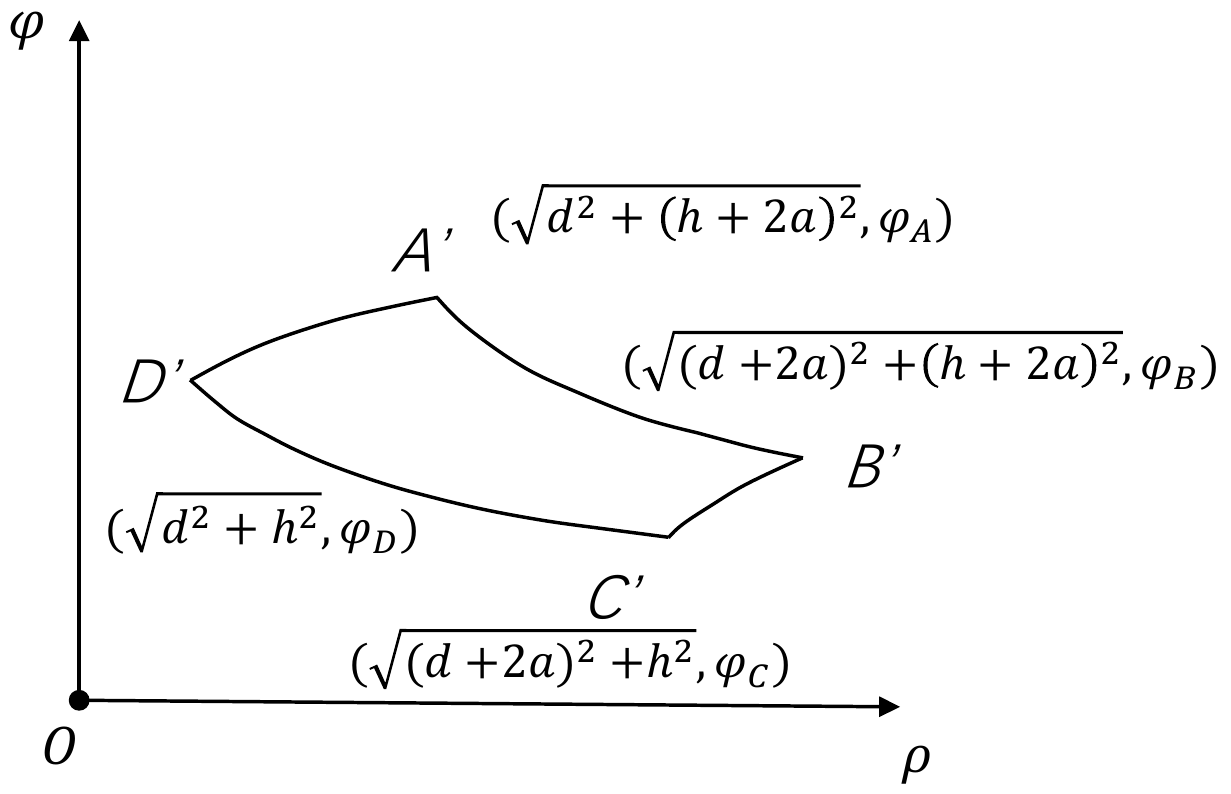}
         \caption{Polar BEV feature map}
         \label{fig:supp_polar_coord}
     \end{subfigure}
     \caption{Object on feature maps with different coordinates.}
     \label{fig:supp_coord}
\end{figure*}

We consider the bounding box of an object as a square with length $2a$ ($a>0)$ and denote the object by $ABCD$ and $A’B’C’D’$  in the Cartesian and Polar BEV feature map, respectively (Figure \ref{fig:supp_cartesian_coord} and \ref{fig:supp_polar_coord}) .The occupied area of the object in the Polar BEV feature map is denoted by $S$. Note that the coordinate of $D$ is $(d, h)$, where $d\in[1,50], h\in[1, 50]$, denoting the detection range.

Regarding $d$ as the only variable, we could  prove $S(d)$ is monotonically decreasing functioned with $d$ by calculating $S'(d)$. 

First, the functions in  Polar BEV feature map for the curve $D'A'$, $A'B'$, $B'C'$, and $C'D'$ are
\begin{equation}
\begin{split}
    &\varphi_{D^{\prime}A^{\prime}}(\rho,d)=\arctan\frac{\sqrt{\rho^2-d^2}}{d},\\
    &\rho\in[\sqrt{d^2+h^2},\sqrt{d^2+(h+2a)^2} ],
\end{split}
\end{equation}
\begin{equation}
\begin{split}
    &\varphi_{A^{\prime}B^{\prime}}(\rho,d)=\arctan\frac{h+2a}{\sqrt{\rho^2-(h+2a)^2}},\\
    &\rho\in[\sqrt{d^2+(h+2a)^2},\sqrt{(d+2a)^2+(h+2a)^2}
\end{split}
\end{equation}
\begin{equation}
\begin{split}
    &\varphi_{B^{\prime}C^{\prime}}(\rho,d)=\arctan\frac{\sqrt{\rho^2-(d+2a)^2}}{d+2a},\\
    &\rho\in[\sqrt{(d+2a)^2+h^2},\sqrt{(d+2a)^2+(h+2a)^2}]
\end{split}
\end{equation}

\begin{equation}
\begin{split}
    &\varphi_{C^{\prime}D^{\prime}}(\rho,d)=\arctan\frac{h}{\sqrt{\rho^2-h^2}},\\
    &\rho\in[\sqrt{d^2+h^2},\sqrt{(d+2a)^2+h^2}]
\end{split}
\end{equation}

Thus, the area of curve $D^{\prime}A^{\prime}B^{\prime}C^{\prime}$ is
\begin{align*}\tag{5}
S(d)&=\left(\int_{\sqrt{d^2+h^2}}^{\sqrt{d^2+(h+2a)^2}} \varphi_{D^{\prime}A^{\prime}}(\rho,d) d\rho +  \right.\\ &\left. \int_{\sqrt{d^2+(h+2a)^2}}^{\sqrt{(d+2a)^2+(h+2a)^2}} \varphi_{A^{\prime}B^{\prime}}(\rho,d) d\rho\right)\\
&-\left(\int_{\sqrt{(d+2a)^2+h^2}}^{\sqrt{(d+2a)^2+(h+2a)^2}} \varphi_{B^{\prime}C^{\prime}}(\rho,d) d\rho+ \right.\\ &\left.
\int_{\sqrt{d^2+h^2}}^{\sqrt{(d+2a)^2+h^2}} \varphi_{C^{\prime}D^{\prime}}(\rho,d) d\rho\right) \\
&=\left(\int_{\sqrt{d^2+h^2}}^{\sqrt{d^2+(h+2a)^2}} \arctan\frac{\sqrt{\rho^2-d^2}}{d} d\rho + 
\right.\\ &\left.\int_{\sqrt{d^2+(h+2a)^2}}^{\sqrt{(d+2a)^2+(h+2a)^2}} \arctan\frac{h+2a}{\sqrt{\rho^2-(h+2a)^2}} d\rho\right)\\
& -\left(\int_{\sqrt{(d+2a)^2+h^2}}^{\sqrt{(d+2a)^2+(h+2a)^2}} \arctan\frac{\sqrt{\rho^2-(d+2a)^2}}{d+2a} d\rho+\right.\\ &\left.
\int_{\sqrt{d^2+h^2}}^{\sqrt{(d+2a)^2+h^2}} \arctan\frac{h}{\sqrt{\rho^2-h^2}} d\rho\right)
\end{align*}

To proof $S(d)$ is monotonically decreasing functioned with $d$, we are going to show $S^{\prime}(d)<0. (d\in [1,50], h\in [1, 50], a > 0)$

First, we get $S^{\prime}(d)$ as follows:
\begin{align*}\tag{6}
S^{\prime}(d)
&=\left(\int_{\sqrt{d^2+h^2}}^{\sqrt{d^2+(h+2a)^2}}-\frac{1}{\sqrt{\rho^2-d^2}}d\rho+\right.\\ &\left.
\frac{d}{\sqrt{d^2+(h+2a)^2}}\arctan\frac{h+2a}{d}-\right.\\ &\left.
\frac{d}{\sqrt{d^2+h^2}}\arctan\frac{h}{d}\right)\\
& +\left(\frac{d+2a}{\sqrt{(d+2a)^2+(h+2a)^2}}\arctan\frac{h+2a}{d+2a}-\right.\\ &\left.
\frac{d}{\sqrt{d^2+(h+2a)^2}}\arctan\frac{h+2a}{d}\right)\\
& -\left(\int_{\sqrt{(d+2a)^2+h^2}}^{\sqrt{(d+2a)^2+(h+2a)^2}}-\frac{1}{\sqrt{\rho^2-(d+2a)^2}}d\rho+\right.\\ &\left.
\frac{d+2a}{\sqrt{(d+2a)^2+(h+2a)^2}}\arctan\frac{h+2a}{d+2a}-\right.\\ &\left. 
\frac{d+2a}{\sqrt{(d+2a)^2+h^2}}\arctan\frac{h}{d+2a}\right)\\
&-\left(\frac{d+2a}{\sqrt{(d+2a)^2+h^2}}\arctan\frac{h}{d+2a}-\right.\\ &\left.
\frac{d}{\sqrt{d^2+h^2}}\arctan\frac{h}{d}\right)\\
&=\int_{\sqrt{d^2+h^2}}^{\sqrt{d^2+(h+2a)^2}} -\frac{1}{\sqrt{\rho^2-d^2}}d\rho-\\
&\int_{\sqrt{(d+2a)^2+h^2}}^{\sqrt{(d+2a)^2+(h+2a)^2}}-
\frac{1}{\sqrt{\rho^2-(d+2a)^2}}d\rho\\
& =\ln\frac{\sqrt{(d+2a)^2+(h+2a)^2}+h+2a}{\sqrt{(d+2a)^2+h^2}+h}-\\
&\ln\frac{\sqrt{d^2+(h+2a)^2}+h+2a}{\sqrt{d^2+h^2}+h}\end{align*}
Take $f(x)=\frac{\sqrt{x^2+(h+2a)^2}+h+2a}{\sqrt{x^2+h^2}+h}, x\in [d,d+2a]$, we have
\begin{align*}\tag{7}
f^{\prime}(x)&=\left(\frac{x}{\sqrt{x^2+(h+2a)^2}}(\sqrt{x^2+h^2}+h)-\right.\\ &\left.
\frac{x}{\sqrt{x^2+h^2}}(\sqrt{x^2+(h+2a)^2}+h+2a)\right)\\
&\cdot(\sqrt{x^2+h^2}+h)^2 \\
&=\frac{x}{(\sqrt{x^2+h^2}+h)^2\sqrt{x^2+(h+2a)^2}\sqrt{x^2+h^2}}\\
&\cdot\left[\sqrt{x^2+h^2}(\sqrt{x^2+h^2}+h)-\right.\\ &\left.\sqrt{x^2+(h+2a)^2}(\sqrt{x^2+(h+2a)^2}+h+2a)\right] \\
&=\frac{x}{(\sqrt{x^2+h^2}+h)^2\sqrt{x^2+(h+2a)^2}\sqrt{x^2+h^2}}\\
&\cdot\left[q(h)-q(h+2a)\right] < 0\end{align*}
where, $q(t)=\sqrt{x^2+t^2}(\sqrt{x^2+t^2}+t)$ is a monotonically increasing function in the domain $[h,h+2a]$
Obviously, $f(x)$ is a monotonically decreasing function with positive values in the domain $[d,d+2a]$. And $g(x)=\ln(x)$ is a monotonically increasing function in the positive field. Thus, $g(f(x)$ is monotonically decreasing functioned with $x$ in the domain $[d,d+2a]$, which means $g(f(d+2a))< g(f(d))$. Therefore, we have
\begin{align*}\tag{8}
S^{\prime}(d)
&=\ln\frac{\sqrt{(d+2a)^2+(h+2a)^2}+h+2a}{\sqrt{(d+2a)^2+h^2}+h} \\
&-\ln\frac{\sqrt{d^2+(h+2a)^2}+h+2a}{\sqrt{d^2+h^2}+h} \\
&=g(f(d+2a))-g(f(d))< 0
\end{align*}

So the occupied area of the object decreases when $d$ increases, and symmetrically, we could obtain the same conclusion for $h$. When objects move far away from the ego-vehicle in the radial direction, both $d$ and $h$ increase, thus $S$ will decrease too. For a generic object with arbitrary orientation, we could consider that it is formed by multiple small squares and the same conclusion can be drawn.


\end{document}